\documentclass[10pt,twocolumn,letterpaper]{article}

\usepackage[pagenumbers]{cvpr} 

\usepackage[utf8]{inputenc} 
\usepackage[T1]{fontenc}    
\usepackage{hyperref}       
\usepackage{url}            
\usepackage{booktabs}       
\usepackage{amsfonts}       
\usepackage{nicefrac}       
\usepackage{microtype}      
\usepackage{xcolor}         
\usepackage{graphicx}
\usepackage{amsmath}
\usepackage{amssymb}
\usepackage{booktabs}
\usepackage{algorithm}
\usepackage{algorithmic}
\usepackage{amsthm}
\usepackage{diagbox}
\usepackage{multirow}

\usepackage[capitalize]{cleveref}
\usepackage{hyperref}
\crefname{section}{Sec.}{Secs.}
\Crefname{section}{Section}{Sections}
\Crefname{table}{Table}{Tables}
\crefname{table}{Tab.}{Tabs.}

\newtheorem{theorem}{Theorem}[section]
\newtheorem{lemma}{Lemma}[section]

\def\cvprPaperID{4062} 
\def\confName{CVPR}
\def\confYear{2023}

\begin{document}

\title{Mutual Information Regularization for Vertical Federated Learning}
\author{Tianyuan Zou \quad Yang Liu \quad Ya-Qin Zhang\\
Institute for AI Industry Research, Tsinghua University\\
Beijing, China\\
{\tt\small zty22@mails.tsinghua.edu.cn, liuy03@air.tsinghua.edu.cn, zhangyaqin@tsinghua.edu.cn}
}

\maketitle

\begin{abstract}
Vertical Federated Learning (VFL) is widely utilized in real-world applications to enable collaborative learning while protecting data privacy and safety.
However, previous works show that parties without labels (\textit{passive parties}) in VFL can infer the sensitive label information owned by the party with labels (\textit{active party}), or execute backdoor attacks to VFL. Meanwhile, active party can also infer sensitive feature information from passive party. 
All these pose new privacy and security challenges to VFL systems.
We propose a new general defense method which limits the mutual information between private raw data, including both features and labels, and intermediate outputs to achieve a better trade-off between model utility and privacy. We term this defense \textbf{M}utual \textbf{I}nformation Regularization \textbf{D}efense (MID). We theoretically and experimentally testify the effectiveness of our MID method in defending existing attacks in VFL, including label inference attacks, backdoor attacks and feature reconstruction attacks.
\end{abstract}

\section{Introduction}
\begin{figure}[!tb]
    \centering
    \begin{subfigure}{0.99\linewidth}
      \includegraphics[width=1\linewidth]{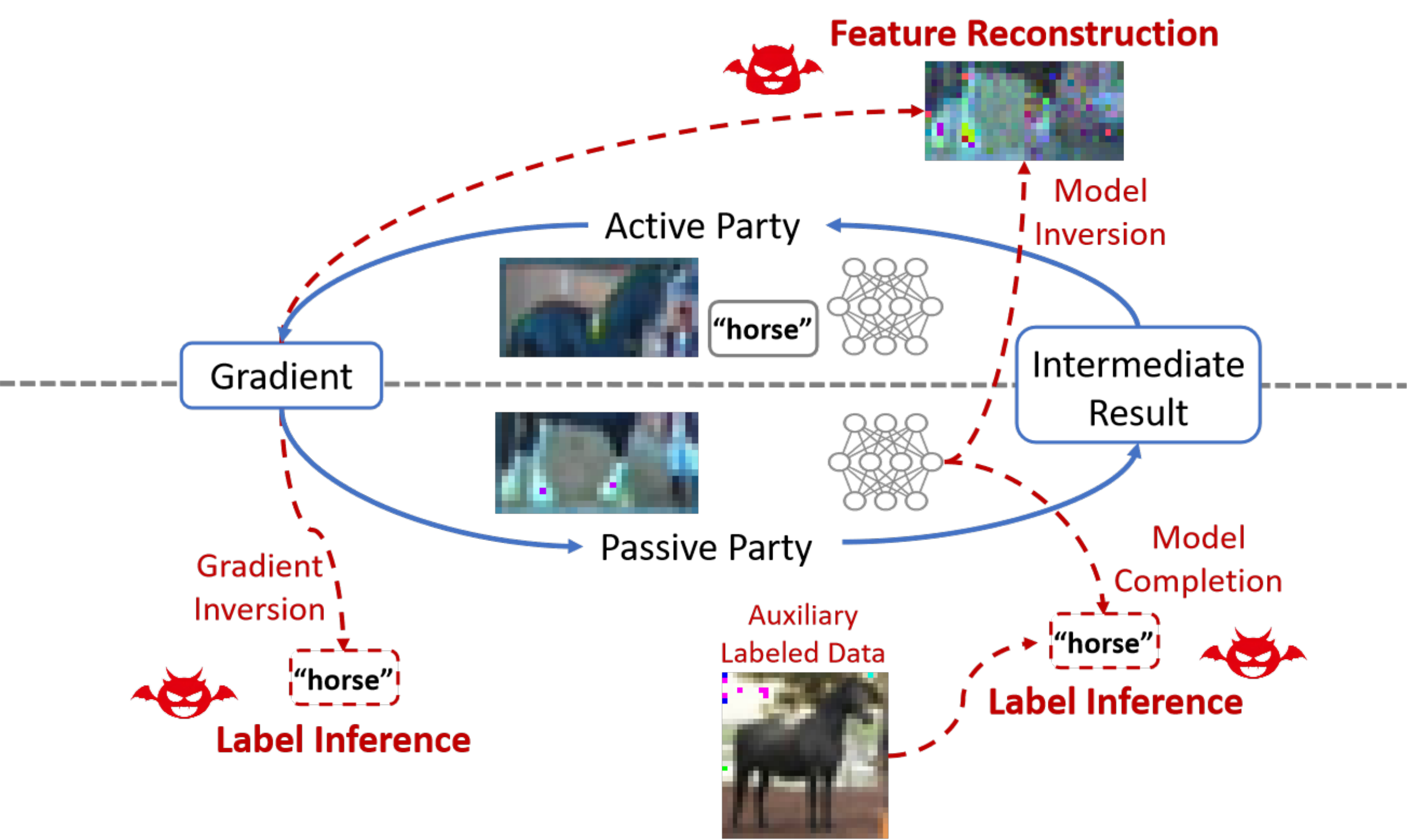}
      \caption{Label Inference Attacks\cite{li2021label,fu2021label,zou2022defending} and Feature Reconstruction Attacks\cite{jin2021cafe}}
      \label{fig:reconstruction_attacks}
    \end{subfigure}
    \\ [1em]
    \begin{subfigure}{0.99\linewidth}
      \includegraphics[width=1\linewidth]{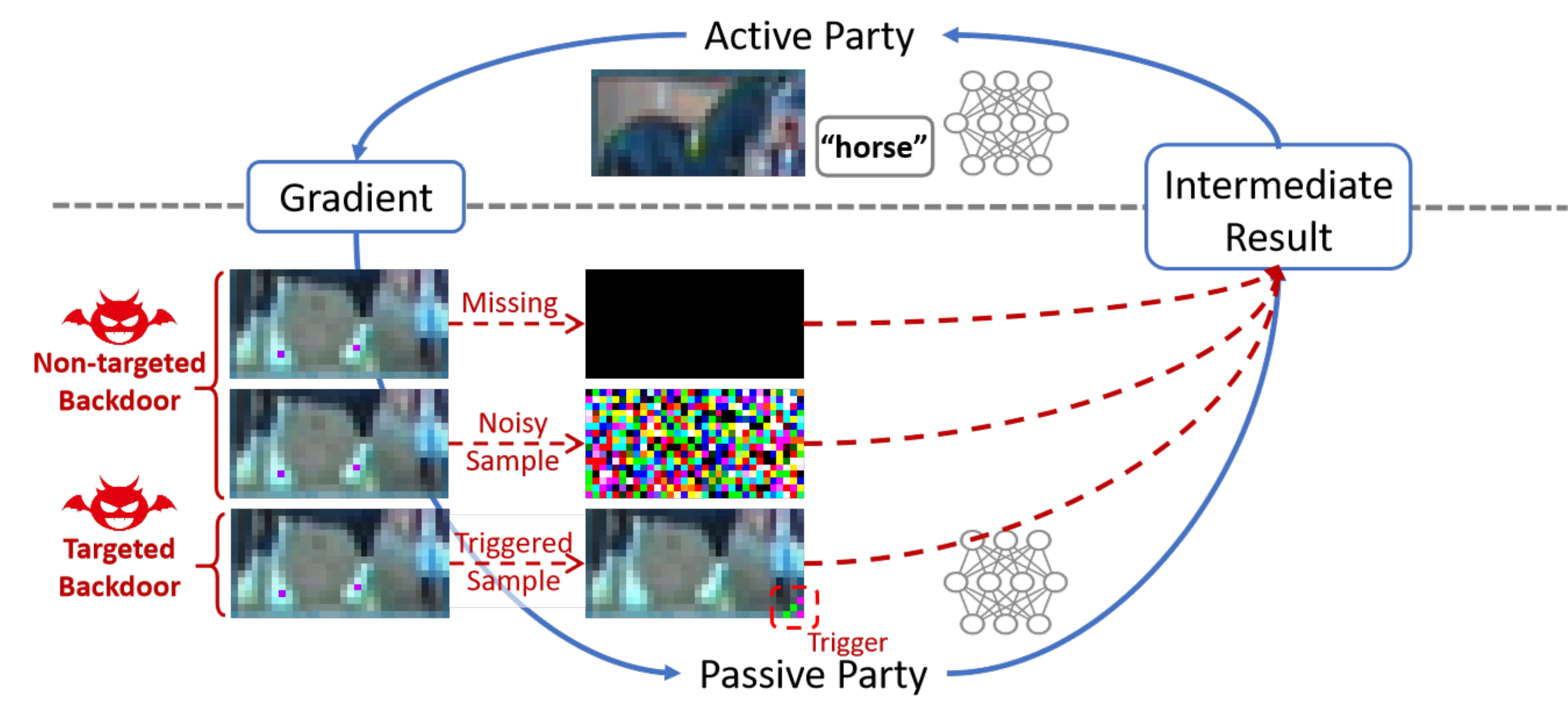}
      \caption{Targeted\cite{zou2022defending} and Non-targeted Backdoor\cite{liu2021rvfr} Attacks}
      \label{fig:backdoor_attacks}
    \end{subfigure}
    \caption{Demonstration of different attacks in $2$-party VFL.}
    \label{fig:attack_demonstrations}
\end{figure}

Federated Learning (FL) \cite{McMahan2016fl} was first proposed to train cross-device machine learning models and protect data privacy simultaneously which can be also regarded as horizontal FL (HFL) \cite{yang2019federatedML} as data are partitioned horizontally in the database by sample. Another kind of FL framework is vertical FL (VFL) \cite{yang2019federatedML,Hu2019FDMLAC,Liu2019ACE,liu2018ftl,cheng2021secureboost,HE2020grouptransfer} where data are partitioned by feature, which means each participant owns a portion of the data features of each sample. This framework is consistent with several real-world situations. For example, a bank and an E-commerce company each obtains some features of the same group of users and they collaboratively train a model for preference prediction. Similar to HFL, participants in VFL aim to collaboratively train a shared model on the premise of keeping their local private data safe by communicating privacy-preserving intermediate results. As shown in \cref{fig:attack_demonstrations}, in basic VFL framework with $2$ parties, local data and local model of each party are kept locally while local intermediate results and gradient information are transmitted between an active and a passive party. 
To attack this basic framework, recent studies\cite{zhu2019deep,zhao2020idlg,yin2021see, zou2022defending} have explored data reconstruction attacks by exploiting the intermediate results exchanged, as well as backdoor attacks by manipulating the input data. As for data reconstruction attacks, both \textbf{\textit{label inference attacks}} \cite{li2021label,fu2021label,zou2022defending} and \textbf{\textit{feature reconstruction attacks}} \cite{jin2021cafe,luo2021feature} have been proposed.  
As for backdoor attacks, malicious passive parties can modify the shared model for their own purpose by adding a trigger to a few of the attacker's local samples in a \textbf{\textit{targeted backdoor attack}} \cite{zou2022defending}, or hurt the overall model utility by adding noise or failing to transmit intermediate results in \textbf{\textit{non-targeted backdoor attacks}}\cite{liu2021rvfr}. We summarize these attacks in \cref{fig:attack_demonstrations}. 

To mitigate these threats, various defense methods can be applied, including Adding Noise \cite{dwork2006DP,Bagdasaryan2018backdoor,xie2020dba}, Gradient Sparsification (GS) \cite{lin2017deep} and Discrete Gradients (DG) \cite{fu2021label}. However, these defense methods suffer from accuracy drop for the main task. There are also specific defense methods for targeted scenarios, such as  MARVELL\cite{li2021label} to defend label leakage in binary classification , Confusional AutoEncoder (CAE) \cite{zou2022defending} to defend label leakage by model inversion attacks, RVFR\cite{liu2021rvfr} to defend robustness-related attacks such as missing features and adversarial input attacks. 
However these defense scenarios are task-specific. 

In this work, we observe that the root cause for data attacks by either active or passive party lies in the fundamental dependency between the local model at a passive party and the label or local features. Therefore we proposed a new general defense method that aims to defend existing attacks from the perspective of information theory. Specifically, we design a \textit{\textbf{M}utual \textbf{I}nformation Regularization \textbf{D}efense} (MID) for restricting the level of information about local data contained in exchanged intermediate outputs. We perform extensive experiments which demonstrate that MID is very effective in defending all kinds of data reconstruction attacks and backdoor attacks compared with existing defense methods. 
Moreover, we provide theoretical guarantee for model robustness with MID under VFL scenario. 

In summary, our contributions are:
\begin{itemize}
\item We propose a new general defense method for VFL, \textit{\textbf{M}utual \textbf{I}nformation Regularization \textbf{D}efense} (MID), which regularizes the information dependency between parties' local sensitive data and exchanged intermediate outputs. We show theoretically that MID is effective in preventing information leakage from exposed intermediate outputs and improving model robustness to defend against backdoor attacks.
\item We perform comprehensive experimental evaluations and show that with proper design of information bottleneck, MID is a promising universal defense method that achieves better utility-privacy trade-off than other general defense methods for various feature reconstruction attacks, label inference attacks and backdoor attacks. 
\end{itemize}

\section{Related Work}


Federated Learning (FL) \cite{mcmahan2016federated,yang2019federatedbook,yang2019federatedML} is a novel machine learning paradigm in which participants collaboratively train a machine learning model without centralizing each parties' local data. FL can be further categorized into horizontal federated learning (HFL) where data are partitioned by samples, and vertical federated learning (VFL) where data are partitioned by features \cite{yang2019federatedML}. 
VFL \cite{liu2019communication,cheng2021secureboost,jiang2022signds} is commonly used in real-world cross-silo applications in finance and advertising\cite{cheng2020federated,FATE}.

Existing attacks to VFL protocols are either to reconstruct private data \cite{li2021label,fu2021label,zou2022defending,jiang2022comprehensive} or to hurt model robustness \cite{liu2021rvfr,liu2020backdoor,zou2022defending,liu2021rvfr,pang2022attacking}. 
For data reconstruction attacks, the target of these attacks is either private labels or private features. Label inference attacks can be performed using sample-level gradients (SLI)\cite{li2021label,fu2021label}, or batch-level gradients (BLI) \cite{zou2022defending}, or trained local models \cite{fu2021label}. Reconstruction of private features also pose great threat to data safety of VFL system. Most related works focus on simple models including logistic regression \cite{luo2021feature,hu2022vertical,weng2020privacy} and tree\cite{luo2021feature}. While for neural networks (NN), recovering image data\cite{jin2021cafe} or tabular data\cite{luo2021feature} can be done by model inversion under white-box setting, and for black-box setting, prior information about data is required\cite{jiang2022comprehensive} or the targeted features are limited to binary values\cite{peng2022binary}. In addition, passive parties can launch backdoor attacks by assigning specific label to triggered samples \cite{zou2022defending}(\textit{targeted} backdoor), or by adding noise to some randomly selected samples or by adding missing features to harm the model utility \cite{goodfellow2014explaining,liu2021rvfr}(\textit{non-targeted} backdoor). 

For defense, cryptographic techniques like Homomorphic Encryption (HE) or Secure Multi-Party Computation (MPC)\cite{yang2019federatedML} have been proposed to protect in-transit messages. However, since they do not protect learned results, VFL with such protections still opens doors to attacks that only exploit trained model results or malicious backdoor\cite{zou2022defending}. Some other general defense strategies focus on reducing information by adding noise \cite{dwork2006DP,li2021label,fu2021label}, Gradient Discretization \cite{dryden2016communication,fu2021label}, Gradient Sparsification \cite{aji2017sparse} and Gradient Compression \cite{lin2017deep}, or combined \cite{shokri2015privacy,fu2021label}. These methods suffer from utility losses. Other emerging defense methods targets to specific attacks or scenarios, such as data augmentation \cite{gao2021privacy} or disguising labels\cite{zou2022defending,jin2021cafe} to defend against gradient inversion attacks, MARVELL\cite{li2021label} to defend against label inference in binary classification tasks, RVFR\cite{liu2021rvfr} to defend against backdoor attacks. 
Mutual Information has been explored as an effective regularization to machine learning models to improve the robustness of model against malicious attacks in the past \cite{alemi2016deep, wang2020infobert,wang2021improving} but has never been explored in VFL setting before. 

\section{Problem Definition}
\subsection{Vertical Federated Learning Setting} \label{sec:vfl_setting}

Under a typical VFL system, $K$ data owners together obtain a dataset of $N$ samples $\mathcal{D} = \{\mathbf{x}_i, y_i\}_{i=1}^{N}$ with each participant $k$ holding a portion of the features $X^k= \{\mathbf{x}^k_i\}_{i=1}^{N}$ and only one party controls the label information $Y=\{y_i\}_{i=1}^{N}$. We refer this party as the \textit{active party}  and other parties as the \textit{passive parties}. Without loss of generality, we assume party $K$ is the active party, and other parties are passive parties. In VFL, each party $k$ adopts a \textit{local model} $G^k$ with model parameters $\theta_k$. Note that $G^k$ can adopt various kinds of model, like logistic regression, tree, support vector machine, neural network, etc. With the local model and data, each participant $k$ calculates its local output $H^k=\{H^k_i\}_{i=1}^{N}=\{G^k(\mathbf{x}_i^k,\theta_k)\}_{i=1}^{N}=G^k(X^k,\theta_k)$ and sends them to the active party for loss calculation. Therefore, 
the overall objective for VFL is formulated as:
\begin{equation}\label{eq:loss}
    \begin{split}
        \min_{\Theta} \mathcal{L}(\Theta; \mathcal{D}) &\triangleq \frac{1}{N}\sum^N_{i=1}\ell(\mathcal{S}(H_i^1,\dots,H_i^K),y_i) 
    \end{split}
\end{equation}
where $\Theta={[{\mathbf{\theta}_{1}};\dots; {\mathbf{\theta}_{K}}]}$ are training parameters, $\mathcal{S}$ denotes a \textit{global model} which can be either a prediction function or a model with trainable parameters, and $\ell$ denotes a loss function, such as a cross entropy loss. 
To perform training with back propagation, 
active party performs gradient computation with received $H^k$ and transmits back $\{\frac{\partial \ell}{\partial H_i}\}_{i=1}^N$ to each party. See \cref{alg:VFL_mid_active} for a complete algorithm.

To further protect transmitted sample-level information, cryptographic techniques such as Homomorphic Encryption (HE)
can be applied \cite{yang2019federatedML} and gradient is calculated under encryption while a coordinator is introduced to the VFL system for distributing encryption keys and decryption. Under HE-protected VFL, sample-level gradient information is protect while batch-level gradient information is revealed.

\begin{algorithm}[H] 
\caption{A VFL framework with and without MID (at active party)}
\textbf{Input}: Learning rate $\eta$; MID hyper-parameter $\lambda$\\
\textbf{Output}: Model parameters $\theta_1,\theta_2,\dots,\theta_K$

\begin{algorithmic}[1] \label{alg:VFL_mid_active}
\STATE Party 1,2,\dots,$K$, initialize $\theta_1$, $\theta_2$, ... $\theta_K$; \\
\FOR{each iteration j=1,2, ...}
\STATE Randomly sample  $S \subset [N]$;
\FOR{each  party $k$ in parallel}
\STATE Computes $\{H_i^k\}_{i \in S}$;
\STATE Sends $\{H_i^k\}_{i \in S}$ to party $K$;
\ENDFOR
\IF{MID is applied} 
\STATE Active party computes $Z_i^k=\mathcal{M_{VIB}}(H_i^k)$ and loss $\ell$ using \cref{eq:mid_loss_multi};
\STATE Active party computes $\{\frac{\partial \ell}{\partial Z_i^k}\}_{i \in S}$ and updates $\mathcal{M_{VIB}}$;
\ELSE
\STATE Active party computes loss $\ell$ using \cref{eq:loss};
\ENDIF
\STATE Active party sends $\{\frac{\partial \ell}{\partial H_i}\}_{i \in S}$ to all other parties;
\FOR{each party k=1,2,\dots,K in parallel}
\STATE Computes $\nabla_k\ell= \frac{\partial \ell}{\partial H_i}\frac{\partial H_i^k}{\partial \theta_k}$;
\STATE Updates $\theta^{j+1}_k = \theta^{j}_k - \eta \nabla_k\ell$;
\ENDFOR
\ENDFOR
\end{algorithmic}
\end{algorithm}

To simplify our discussion,
we first consider a VFL system with $1$ active party and $1$ passive party only, whose input spaces are $X^a$ and $X^p$ respectively. The training objective under this setting can be written as:
\begin{equation}\label{eq:original_loss}
    \mathcal{L} = \ell(\hat{Y},Y) = \ell(\mathcal{S}(H^a,H^p),Y)
\end{equation}
where $H^a,H^p$ are the intermediate local outputs of active party and passive party respectively and $\hat{Y}$ denotes the predicted labels. 
Multi-party scenario can be easily extended and will be studied in the following sections.

\subsection{Attacks}\label{section:all_attacks}

\textbf{Label Inference Attacks.}
In label inference attacks, passive parties try to steal the private labels from the active party. Multiple routes can be taken to complete these attacks: 
    \textit{\textbf{Model Completion attack}} (MC)\cite{fu2021label} infers label by completing the local model with an additional layer and fine-tuning the whole model using auxiliary labeled data. Depending on whether the attacker updates its local model actively to infer more information, MC attack can be separated into \textit{active MC attack} (AMC) and \textit{passive MC attack} (PMC).
    \textit{\textbf{Sample-level Label Inference attack}} (SLI)\cite{li2021label,fu2021label} assumes sample-level gradient information is exposed to the attacker. \textit{Direct Label Inference attack} (DLI)\cite{fu2021label,li2021label} exploits the fact that sample-level gradient $\frac{\partial \ell}{\partial H_i^p}$ exhibits a different sign value on the label position when a global softmax function $\mathcal{S}$ is applied. Assuming the gradient of one random positive sample is known, \textit{Direction Scoring attack}  (DS)\cite{li2021label} exploits the cosine similarity between each gradient pairs to cluster each sample into positive or negative class.
    \textit{\textbf{Batch-level Label Inference attack}} (BLI)\cite{zou2022defending} assumes only the local batch-level gradient is locally available, such as in the case of VFL with HE-protection, and trains a neural network (NN) model to invert label information from batch-level gradients.



\textbf{Backdoor attacks.} Depending on whether a separate training target exhibits, backdoor attacks can be categorized into \textit{targeted} and \textit{non-targeted} backdoor attacks. 
    \textit{\textbf{Targeted Backdoor attack.}} \textit{Gradient Replacement Backdoor attack}\cite{zou2022defending} is a targeted backdoor where the attacker attempts to assign a previously chosen target label $\tau$ to triggered samples. 
    \textit{\textbf{Non-targeted backdoor attacks}} include \textit{Noisy-sample Backdoor attack} which aims to harm the model utility by adding random noise $\delta \mathbf{x}_{(n)}^p$ to randomly chosen samples to get noisy sample ${\mathbf{x}_i^p}^{\prime}$ and \textit{Missing Backdoor attack}\cite{liu2021rvfr} in which some $H_i^p$ are randomly lost (set to $\mathbf{0}$), equivalent to setting ${\mathbf{x}_i^p}^{\prime}=\mathbf{x}_{(m)}^p$ that satisfies ${H^P}^{\prime}=G^p(\mathbf{x}_{(m)}^p)=\textbf{0}$, through out training process.
We summarize the backdoor dataset ${X^p}^{\prime}={\{{\textbf{x}_i^p}^{\prime}\}}_{i=1}^N$ with:
\begin{align*}
\small
    {\textbf{x}_i^p}^{\prime} \triangleq
    \begin{cases}
    \mathbf{x}_i^p + \delta \mathbf{x}_{(t)}^p & \, \text{triggered sample $i$}\\
    \mathbf{x}_i^p + \delta \mathbf{x}_{(n)}^p & \, \text{noisy sample $i$}\\
    \mathbf{x}_{(m)}^p & \, \text{missing sample $i$}\\
    \mathbf{x}_i^p & \, \text{others}
    \end{cases}
\end{align*}
and the false label set $Y^f={\{y_i^f\}}_{i=1}^N$ with:
\begin{align*}
\small
   {y^f_i} \triangleq
    \begin{cases}
    \tau & \, \text{triggered sample $i$}\\
    \tilde{y_i} \ne y_i & \, \text{noisy/missing sample $i$}\\
    y_i & \, \text{others}
    \end{cases}
\end{align*}
Then, the training goal of a backdoor attacker is:
\begin{equation}\label{eq:backdoor_target}
\small
    \begin{split}
        \min_{\Theta} \mathcal{L}^b(\Theta; \mathcal{D}^{\prime}) &\triangleq \frac{1}{N}\sum^N_{i=1}\ell(\mathcal{S}(H_i^a,{H_i^p}^{\prime}),y_i^f) \\&= \frac{1}{N}\sum^N_{i=1}\ell(\mathcal{S}(G^a(\mathbf{x}_i^a),G^p({\mathbf{x}_i^p}^{\prime})),y_i^f)
    \end{split}
\end{equation}


\textbf{Feature Reconstruction Attacks.} Parties in VFL can also utilize its local data and knowledge to reconstruct private local features belonging to other parties. CAFE \cite{jin2021cafe} provides a possible feature reconstruction method by inverting the parties' local models $G^k$ using neural network under a white-box VFL setting, which means that the active party has knowledge of passive parties' local models $\{G^k\}_{k=1}^{K-1}$.


\section{Mutual Information Regularization}

\begin{figure}[!tb]
    \centering
    \begin{subfigure}{0.99\linewidth}
      \includegraphics[width=1\linewidth]{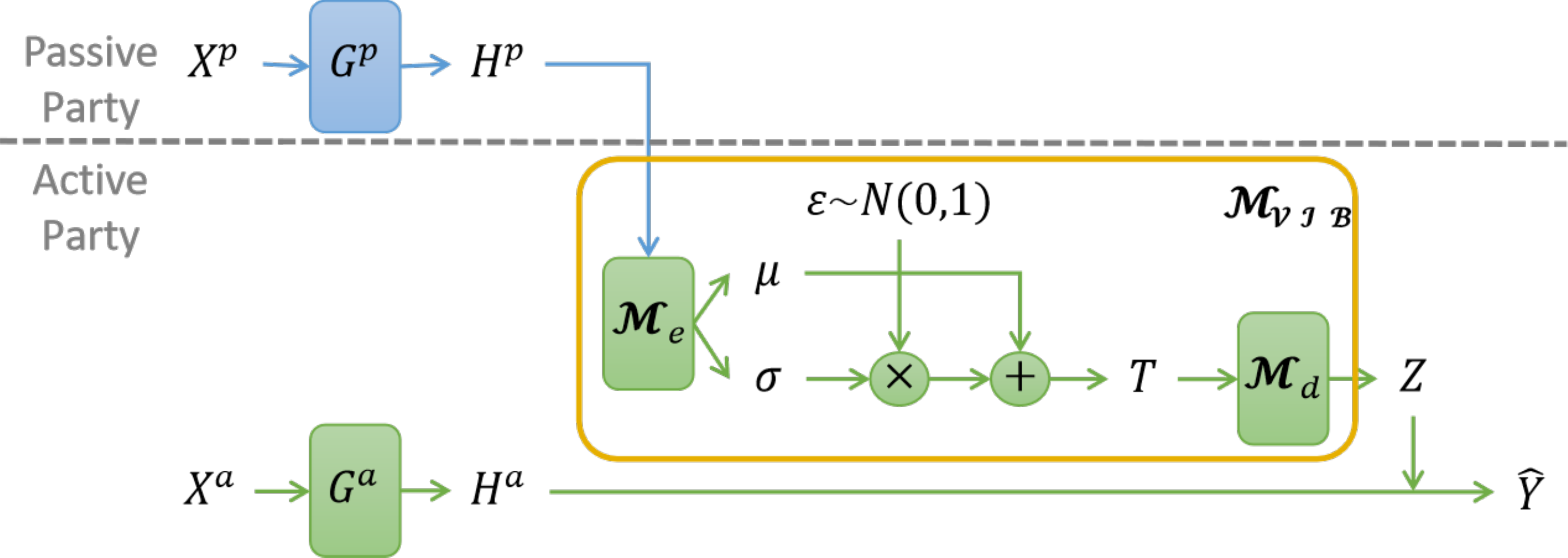}
      \caption{Active Party with MID}
      \label{fig:active_mid_implementation}
    \end{subfigure}
    \\ [1em]
    \begin{subfigure}{0.99\linewidth}
      \includegraphics[width=1\linewidth]{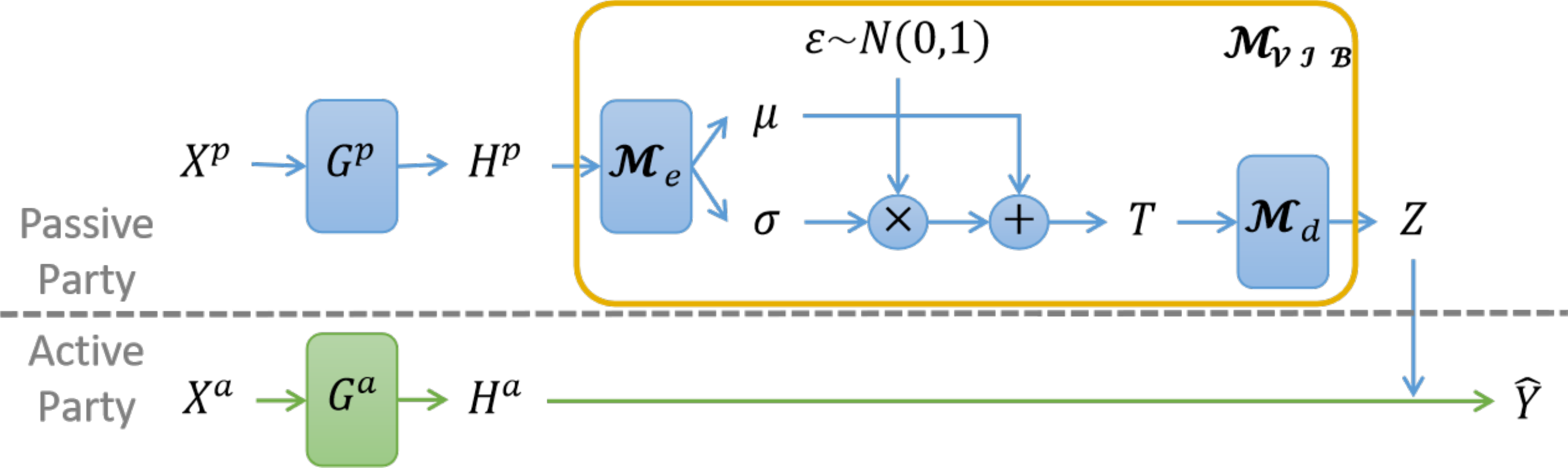}
      \caption{Passive Party with MID}
      \label{fig:passive_mid_implementation}
    \end{subfigure}
    \caption{Demonstration of MID implementation in a $2$-party VFL system. $X^p, X^a$ denotes local data sample set at passive and active party separately.}
    \label{fig:mid_implementation}
\end{figure}

\subsection{Defense Against Label Inference Attacks}\label{sec:mid_theory_label_inference}

In order to prevent all the passive party's attacks in \cref{section:all_attacks} 
, one possible way is for the active party to reduce the dependency of their local models on the label and predicted label. Following works on Information Bottleneck (IB)\cite{tishby1999information,tishby2015deep,alemi2016deep}, we regard the neural network considering $X^p$ as a Markov chain $Y-X^p-H^p-T-Z-\hat{Y}$, where $H^p$ is the original model output, $T$ is a stochastic encoding layer, $Z$ is the new model output which aims to decode $Y$ from $T$ and $\hat{Y}$ is the VFL model prediction. Following the Data Processing Inequality (DPI) theory\cite{beaudry2011intuitive}, $I(H^p,\hat{Y}) \leq I(H^p,T)$. To minimize the mutual information (MI) between $\hat{Y}$ and $H^p$, $I(\hat{Y},H^p)$, following \cite{wang2021improving}, we can replace $I(\hat{Y},H^p)$ with its upper bound $I(H^p,T)$ and the training objective with:

\begin{equation}\label{eq:IB_formualtion}
    \min_{T}\{-I(Y,T)+\lambda I(H^p,T)\}
\end{equation}


Since $I(Y,T)$ is maximized simultaneously as \cref{eq:original_loss} is minimized \cite{boudiaf2020unifying}, we then combine \cref{eq:original_loss,eq:IB_formualtion} to rewrite the loss function for VFL training as the following:
\begin{equation}\label{eq:mid_loss}
\begin{split}
    \mathcal{L} &= \ell(\hat{Y},Y) + \lambda \cdot I(H^p,T)\\
                &= \ell(\mathcal{S}(H^a,Z),Y) + \lambda \cdot I(H^p,T), \lambda \geq 0
\end{split}
\end{equation}
When minimizing $\mathcal{L}$, $I(\hat{Y},Y)$ is maximized to guaranty the model performance while $I(H^p,T)$ is minimized to prevent the passive party from inferring active party's private label information $Y$.
If there exits more than one passive party, the loss function can be generalized as:
\begin{equation}\label{eq:mid_loss_multi}
\small
\begin{split}
    \min_{\Theta} \mathcal{L}(\Theta; \mathcal{D}) \triangleq &\frac{1}{N}\sum^N_{i=1}\ell(\mathcal{S}(Z_i^1,\dots,Z_i^{K-1},H_i^K),y_i)\\
    &+\sum^{K-1}_{k=1} \lambda^k I(H^k,T^k), \lambda^k \geq 0
\end{split}
\end{equation}

Although the idea is straight forward, in reality, it is hard to precisely calculate the mutual information $I(Y,T)$ and $I(H^p,T)$. To overcome this difficulty, we follow the implementation of \textit{Variational Information Bottleneck (VIB)}\cite{alemi2016deep}. The idea is to use parametric modeling to approximate the calculation of those two mutual information value, with an encoder to approximate $I(H^p,T)$ and a decoder to approximate $I(Y,T)$. Reparameterization trick is also applied to make the decoder derivable thus making the backward propagation process possible. The process can be denoted as:
\begin{equation}\label{eq:embedding}
    Z = \mathcal{M_{VIB}}(H^p)
\end{equation}
where $\mathcal{M_{VIB}}$ is the "encoder-decoder" structure with reparameterization trick for derivable guarantee. $\mathcal{M_{VIB}}$ first transmits $H^p$ to the bottleneck layer $T$ which ignores as much detail of $H^p$ as possible but keeps sufficient information about $Y$, and then decodes $Y$ related information from $T$ and outputs $Z$ as the decoded representation of $Y$. Specifically, the encoder $\mathcal{M}_e$ is to estimate the $\mu,\sigma$ for $T$ to achieve $p(t|h^p)=\mathcal{N}(t|\mu,\sigma^2)$ which is needed in the calculation of $I(H^p,T) = \iint p(h^p,t) \log \frac{p(h^p,t)}{p(h^p)p(t)} \, dh^p dt = \iint p(h^p,t) \log \frac{p(t|h^p)}{p(t)} \, dh^p dt$. The stochastic attribute of $T$ lies in the random generation of $T$ according to $\mu,\sigma$, that is $T=\mu+\epsilon \cdot \sigma, \epsilon \sim \mathcal{N}(0,1)$. And the decoder $\mathcal{M}_d$ is a variational approximation to $p(y|t)$ which is needed in the calculation of $I(Y,T) = \iint p(y,t) \log \frac{p(y,t)}{p(y)p(t)} \, dy dt = \iint p(y,t) \log \frac{p(y|t)}{p(y)} \, dy dt$. See \cref{fig:active_mid_implementation} for detailed demonstration. In \cref{fig:mid_implementation}, we use $\mathcal{M}_e, \mathcal{M}_d$ to denote the encoder and decoder inside $\mathcal{M_{VIB}}$, $T$ is the output of reparameterization and $Z$ is the output of $\mathcal{M_{VIB}}$, also is the local model prediction under MID. 

We provide a detailed training algorithm with MID protection in \cref{alg:VFL_mid_active}.
As this defense method is designed from MI perception, we term it \textit{\textbf{M}utual \textbf{I}nformation Regularization \textbf{D}efense (MID)}.
For MID, $\lambda$ is the hyper-parameter that controls the balance between information compression of $H^p$ in $T$ and the representation ability of $T$ according to $Y$. A large $\lambda$ indicates a high compression rate which should result in a better defense ability but may harm the VFL utility at the same time. When $\lambda=0.0$, no information bottleneck regularization is applied but only $\mathcal{M}_e$ and $\mathcal{M}_d$ are added as additional model layers to the VFL system since their existence or absence is regardless of the value of $\lambda$. 




\subsection{Defense Against Backdoor Attacks}\label{sec:mid_theory_backdoor}



\begin{lemma}\label{lemma:for_backdoor_theorem}
When MID is applied, the goal of defending against backdoor attacks is to $\min |I(Y,T)-I(Y,T^{\prime})|$ where $T, T^{\prime}$ is the MID bottleneck representation for the original and the modified local data sample.
\end{lemma}

\begin{proof}
As describe in \cref{section:all_attacks}, in targeted and non-targeted backdoor attacks, the passive attacker aims to achieve \cref{eq:backdoor_target}, making the prediction $\hat{y}^{\prime}$ closer to $y^f_i$ rather than the sample's original label $y_i$. Let $\hat{Y}^{\prime}=\{\hat{y}_i^{\prime}\}_{i=1}^N=\mathcal{S}(H^a,{H^p}^{\prime})$, then $I(\hat{Y}^{\prime},Y^f) \geq I(\hat{Y}^{\prime},Y)$ while $I(\hat{Y},Y^f) \leq I(\hat{Y},Y)$ holds for true for $\hat{Y}=\mathcal{S}(H^a,H^p)$. Therefore, to defend against all these backdoor attacks, the goal is to minimize the change in $I(\hat{Y}^{\prime},Y)$ compared to $I(\hat{Y},Y)$, that is to $\min |I(\hat{Y}^{\prime},Y) -I(\hat{Y},Y)|$. 
With ${H^p}^{\prime}$ converted to $T^{\prime}$ in MID, this is equivalent to 
\begin{align} \label{eq:backdoor_defense_target}
    \min |I(Y,T)-I(Y,T^{\prime})| 
\end{align}
\end{proof}

\begin{theorem}\label{theorem:mid_backdoor}
The performance gap $|I(Y,T)-I(Y,T^{\prime})|$ is bounded by the following:

\begin{equation} \label{eq:infobert_theorem}
\small
\begin{split}
    |I(Y,T)-I(Y,T^{\prime})| &\leq B_1 {\mathcal{|T|}}^{1/2}(I(H^p,T))^{1/2}\\
    &+ B_2 {\mathcal{|T|}}^{3/4}(I(H^p,T))^{1/4}\\ 
    &+ B_3 {\mathcal{|T|}}^{1/2}(I({H^p}^{\prime},T^{\prime}))^{1/2}\\
    &+ B_4 {\mathcal{|T|}}^{3/4}(I({H^p}^{\prime},T^{\prime}))^{1/4} + B_0
\end{split}
\end{equation}
where $B_1=B_2 \log\frac{1}{B_2}, B_2 = \frac{4\sqrt{2 \log2}}{\min_{{h^p} \in \mathcal{H}^p} \{p(h^p)\}}, B_3=B_4 \log\frac{1}{B_4}, B_4 = \frac{4\sqrt{2 \log2}}{\min_{{h^p}^{\prime} \in {\mathcal{H}^p}^{\prime}}\{p({h^p}^{\prime})\}}, B_0 = \log M$ and $M=\sup_{t \in \mathcal{T}}\{M(t)\}$ with $M(t)$ being the number of adversarial representation $t^{\prime} \in {\mathcal{T}}^{\prime}=\mathcal{T}$ that satisfies $||t-t^{\prime}||_2 \leq \epsilon$ given any $\epsilon > 0$.
\end{theorem}

Thus, according to \cref{theorem:mid_backdoor} and \cref{eq:mid_loss}, when the active party applies MID, by improving model robustness, backdoor attacks launched by the passive party is prevented.

\subsection{Defense Against Feature Reconstruction Attacks}\label{sec:mid_theory_data_inference}
When the attacker's target is to recover features, the defending party can also utilize MID to protect its data by adding $\mathcal{M_{VIB}}$ behinds its original local model output $H^p$, generating $Z^p=\mathcal{M_{VIB}}(H^p)$ to further decrease $I(X^p,Z^p)$. With MID, the defender (passive party) is able to defend against feature reconstruction attacks, even for attacks that directly exploits local models such as CAFE\cite{jin2021cafe}. See \cref{fig:passive_mid_implementation} for its implementation. Note that, from the omniscient perspective, the whole model architecture is the same whether the MID defense is implemented in the passive or active party. A detailed training algorithm is provided in \cref{alg:VFL_mid_passive} in the appendix.


\begin{theorem}
When applying MID, passive party is able to protect local private data $X^p$ by minimizing $I(X^p,Z)$. 
\end{theorem}

\begin{proof}

In the case of MID implemented in the passive party, the Markov chain $Y-X^p-H^p-T-Z-\hat{Y}$ can still apply. As the reverse sequence of a Markov chain also forms a Markov chain, according to DPI theory\cite{beaudry2011intuitive}, we have $I(X^p,Z) \leq I(X^p,T) \leq I(H^p,T)$. Since $I(H^p,T)$ is an upper bound of $I(X^p,Z)$, $I(X^p,Z)$ is simultaneously minimized as \cref{eq:IB_formualtion} is achieved. So, the passive party can still apply this objective function for its MID and obtains a stochastic layer $T$ containing all the available knowledge about $Y$ but only the minimum sufficient statistical knowledge about $H^p$ and $X^p$. 
\end{proof}

\section{Experiments}


\subsection{Models and Datasets}\label{subsection:models_and_datasets}
We conduct our experiments on $3$ different datasets: MNIST, CIFAR10 and CIFAR100.
In MNIST dataset\cite{MNISTdataset}, each image sample is evenly split and assigned to each party respectively. A 2-layer MLP model with a 32-neuron layer as the hidden middle layer is used for each party's local model except in CAFE attack which adopts a Convolution-MaxPool-Convolution-MaxPool model structure followed by a 3-layer-FC model as each party's local model following the original work\cite{jin2021cafe}.
In CIFAR10 and CIFAR100 dataset\cite{krizhevsky2009learning}, each image sample is evenly split and assigned to each party respectively. Resnet20 is used for each party's local model in model completion attacks (PMC and AMC)  to be consistent with the original work\cite{fu2021label} and the same model structure for MNIST dataset is applied for these $2$ datasets in CAFE. While for other attacks, Resnet18 is used. 

Through out our experiments, all data from the $3$ datasets are used for multi-class classification tasks. We use the training and testing dataset provided therein. For binary classification tasks, we randomly select $2$ classes and use the belonging data to compose a balanced dataset.

As for the global prediction model $\mathcal{S}$, a global softmax function is used at the active party, except for MC attacks\cite{fu2021label}, DS attack and CAFE attack\cite{jin2021cafe}, which adopts a 4-layer FC model, 1-layer FC model and 1-layer FC model respectively with trainable parameters. Global trainable model is not used for other attacks, namely DLI attack, BLI attack, targeted and non-targeted backdoor attacks, in order to guarantee a stronger attack performance. 

\subsection{Attacks} \label{subsection:experiments_attacks}
We test the effectiveness of MID on $9$ kinds of attacks designed for VFL systems, namely, Passive Model Completion attack (PMC)\cite{fu2021label}, Active Model Completion attack (AMC)\cite{fu2021label}, Direct Label Inference attack (DLI)\cite{li2021label,fu2021label}, Direction Scoring attack (DS)\cite{li2021label}, Batch-level Label Inference attack (BLI)\cite{zou2022defending},
Label Replacement Backdoor attack\cite{zou2022defending},
Noisy-sample attack\cite{liu2021rvfr}, Missing attack\cite{liu2021rvfr} and CAFE\cite{jin2021cafe}. The first $5$ attacks are label inference attacks, the last attack is feature reconstruction attack, and the rest are backdoor attacks.

For MC attacks, we use CIFAR10 dataset with $40$ and $10$ auxiliary labeled data and CIFAR100 dataset with $400$ and $100$ auxiliary labeled data, which means each class of CIFAR10 or CIFAR100 owns $4$ or $1$ auxiliary labeled data belonging to that class.
In BLI attack, we follow the implementation detail in \cite{zou2022defending} which means batch size is set to $2048$. For label replacement backdoor attack, $1\%$ of data samples are randomly selected and marked with trigger while target label $\tau$ is also randomly chosen\cite{zou2022defending}. $1\%$ of data samples are added with noise $\delta \mathbf{x}^p_{(n)} \sim \mathcal{N}(0,2)$ for noisy-sample attack, while $25\%$ of passive model outputs failed to get to the active party, i.e. ${H_i^p}^{\prime}=\mathbf{0}$, for missing attack. For CAFE, we follow the CAFE implementation\cite{jin2021cafe} with default hyper-parameters and use a batch size of $40$ with the number of iterations for feature reconstruction set to $10000$ for MNIST and $20000$ for CIFAR10 and CIFAR100. Notice that the \textit{first FC layer}, of which CAFE first recovers its output and input before recovering the input data sample features, is selected differently depending on whether MID is applied. If MID is applied, the \textit{first FC layer} is the one-layer MID decoder, also the last layer. Otherwise, same as the original paper\cite{jin2021cafe}, there are $2$ more FC layers after the \textit{first FC layer}. 

\subsection{Baseline Defense Methods} \label{subsection:defense_methods}
In our experiments, we evaluate MID with $3$ general defense method: Adding Noise with Gaussian distribution (DP-G) or Laplace distribution (DP-L) and Gradient Sparsification (GS).
We also evaluate DiscreteSGD (DG)\cite{fu2021label} against MC attacks and DLI attack, MARVELL\cite{li2021label} against DS attack which is conducted under binary classification task, Confusional AutoEncoder (CAE)\cite{zou2022defending} against BLI attack and RVFR\cite{liu2021rvfr} against backdoor attacks.

\textbf{Adding Noise.} A Gaussian or Laplacian noise with standard deviation ranging from $5e^{-5}$ to $1.0$ is added to the gradients after they are $2$-norm clipped with $0.2$. Gaussian noise is also added to defend against data reconstruction attack in which gradients are $2$-norm clipped with $3$ with noise of standard deviation ranging from $0.1$ to $10$ added.
\textbf{GS.}\cite{aji2017sparse} Various drop rate ranging from $50.0\%$ to $99.9\%$ is evaluated in the experiment.
\textbf{DG.}\cite{fu2021label} Number of bins for gradient quantification ranging from $3$ to $24$ is evaluated in the experiments.
\textbf{MARVELL.}\cite{li2021label} 
The power constraint hyper-parameter ranging from $0.1$ to $10$ times the norm of gradients is evaluated.
\textbf{CAE.}\cite{zou2022defending} Following the original paper, both encoder and decoder of CAE have the architecture of 2-layer-FC.
Hyper-parameter $\lambda_2$ that controls the confusion level ranging from $0.0$ to $2.0$ is evaluated.
\textbf{RVFR.}\cite{liu2021rvfr}  We evaluate this defense method in backdoor attacks following the default parameter setting of the original paper.
Note that, the forth \textit{server training} stage in RVFR is inapplicable under our VFL setting as no trainable global model exits.

\begin{figure} [!tb]
  \centering
  \begin{subfigure}{0.49\linewidth}
    \includegraphics[width=1\linewidth]{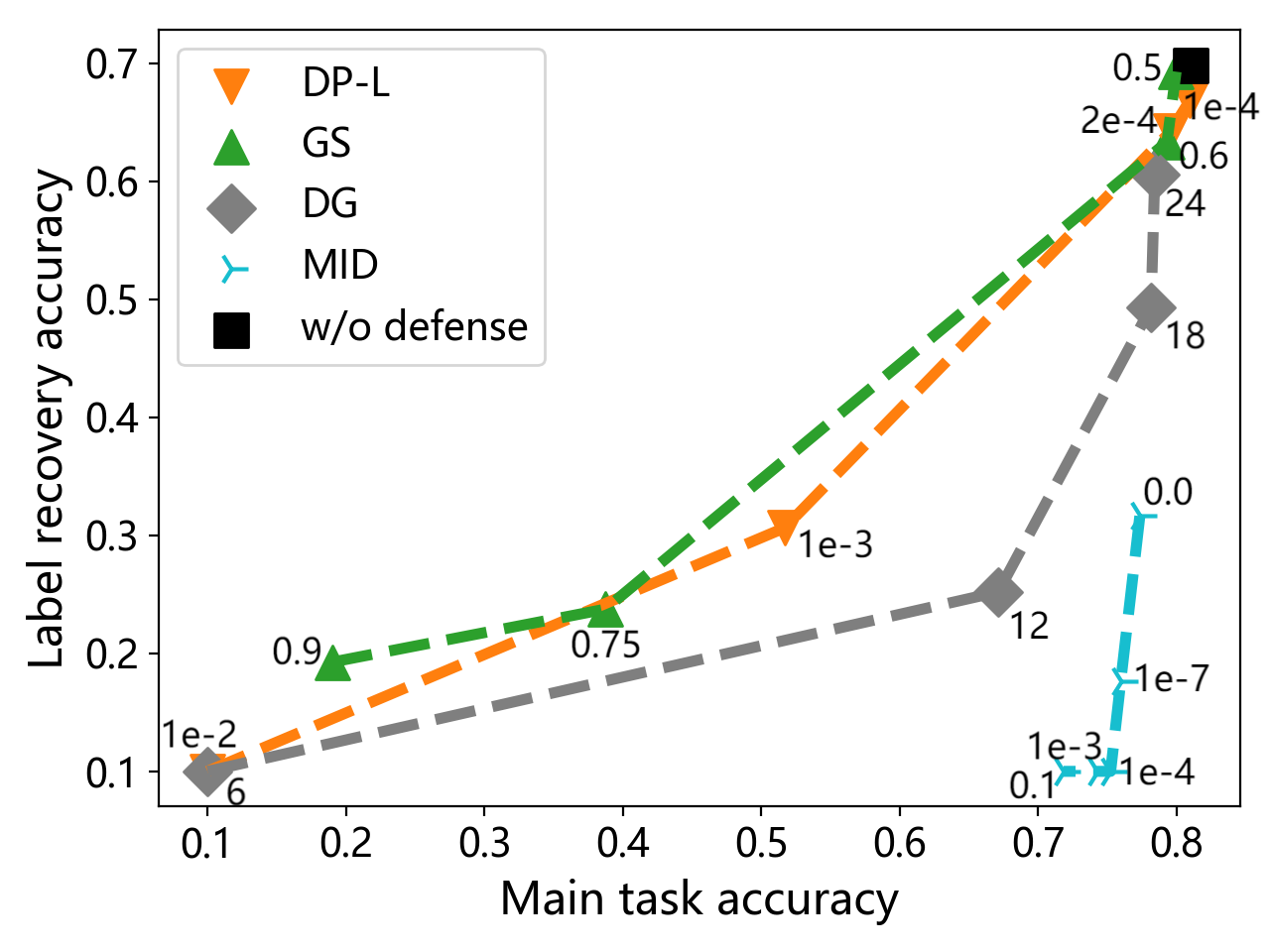}
    \caption{CIFAR10 PMC-40}
    \label{fig:pmc_40_cifar10}
  \end{subfigure}
  \hfill
  \begin{subfigure}{0.49\linewidth}
    \includegraphics[width=1\linewidth]{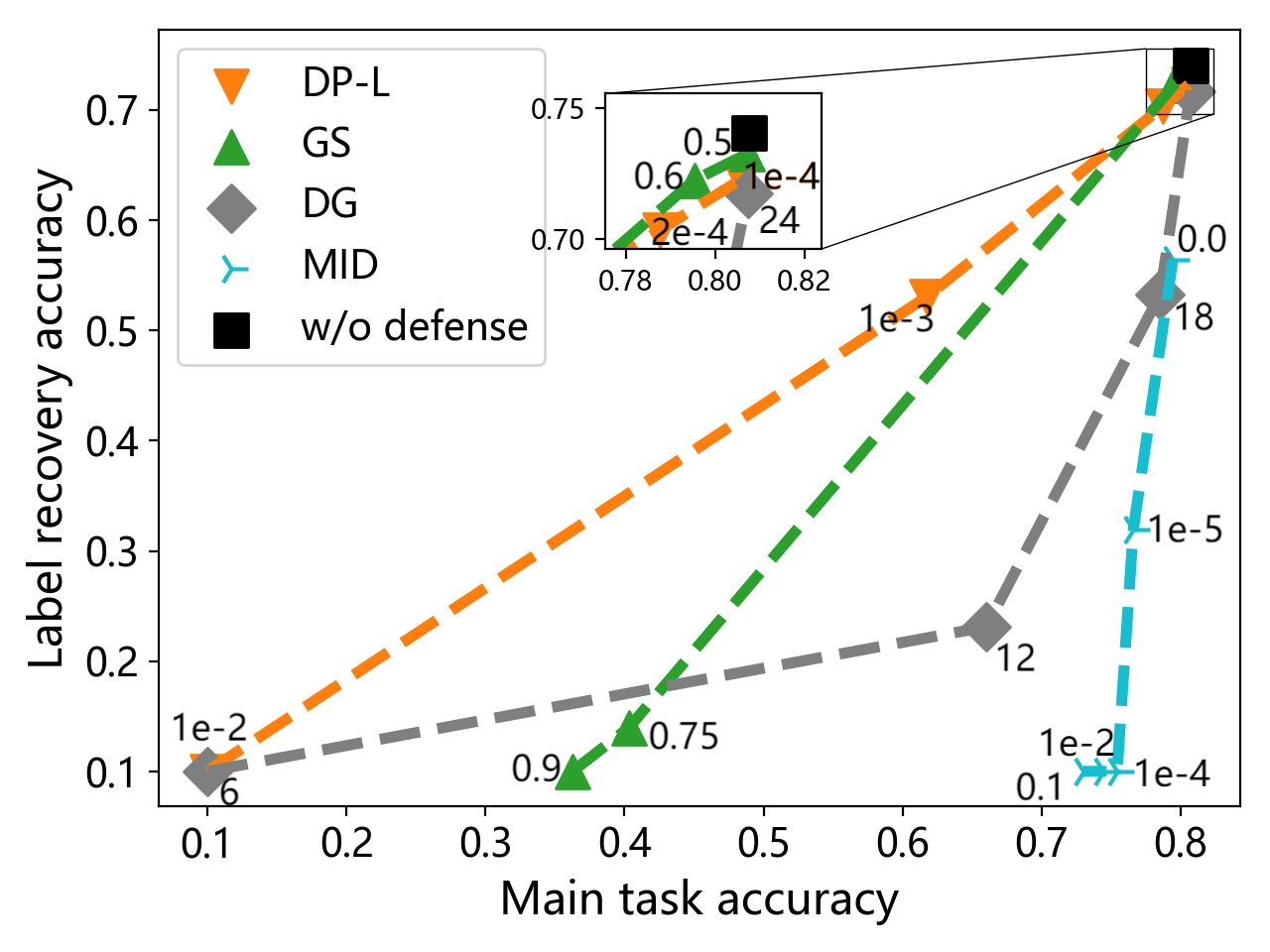}
    \caption{CIFAR10 AMC-40}
    \label{fig:amc_40_cifar10}
  \end{subfigure}

  \begin{subfigure}{0.49\linewidth}
    \includegraphics[width=1\linewidth]{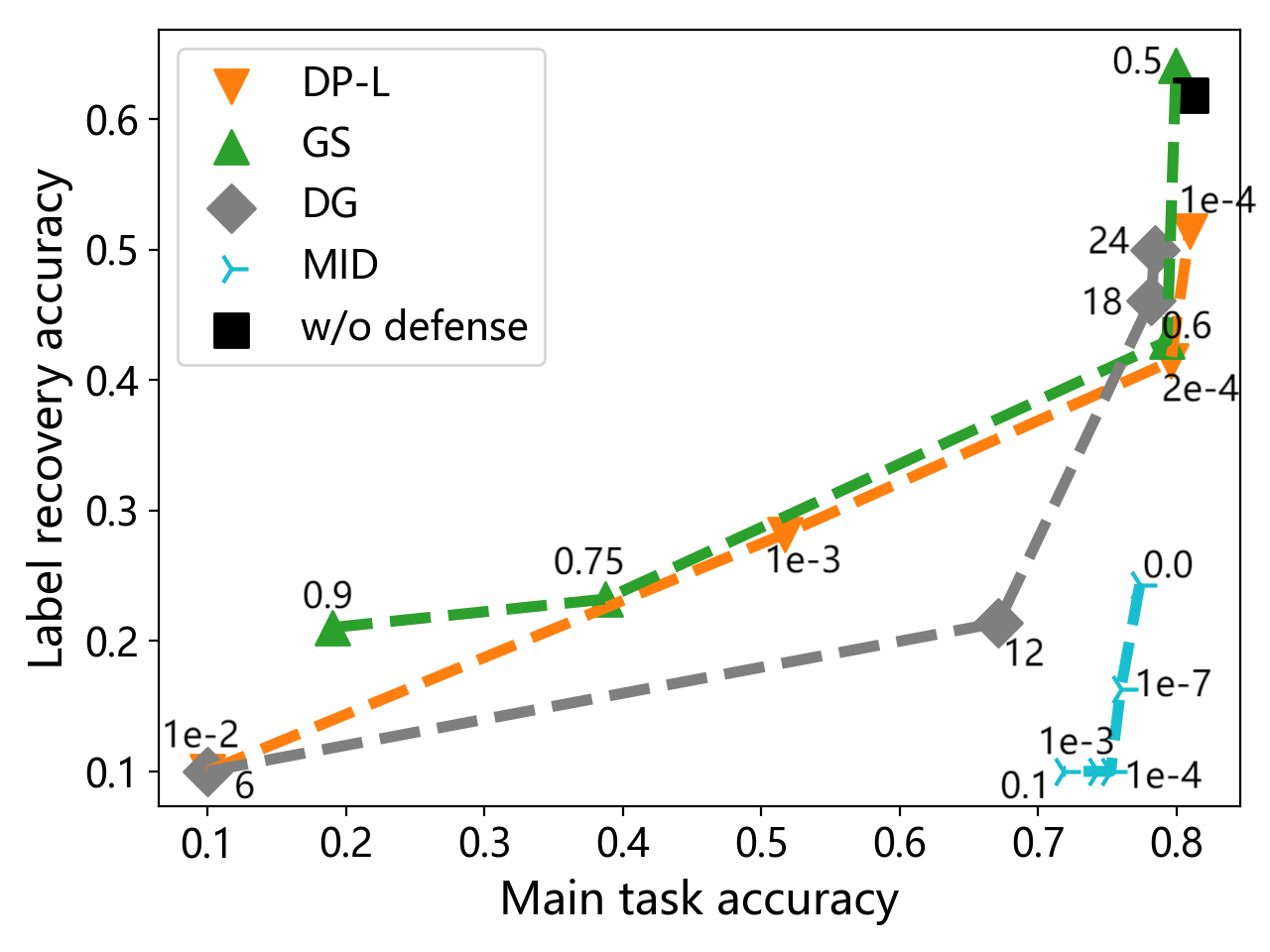}
    \caption{CIFAR10 PMC-10}
    \label{fig:pmc_10_cifar10}
  \end{subfigure}
  \hfill
  \begin{subfigure}{0.49\linewidth}
    \includegraphics[width=1\linewidth]{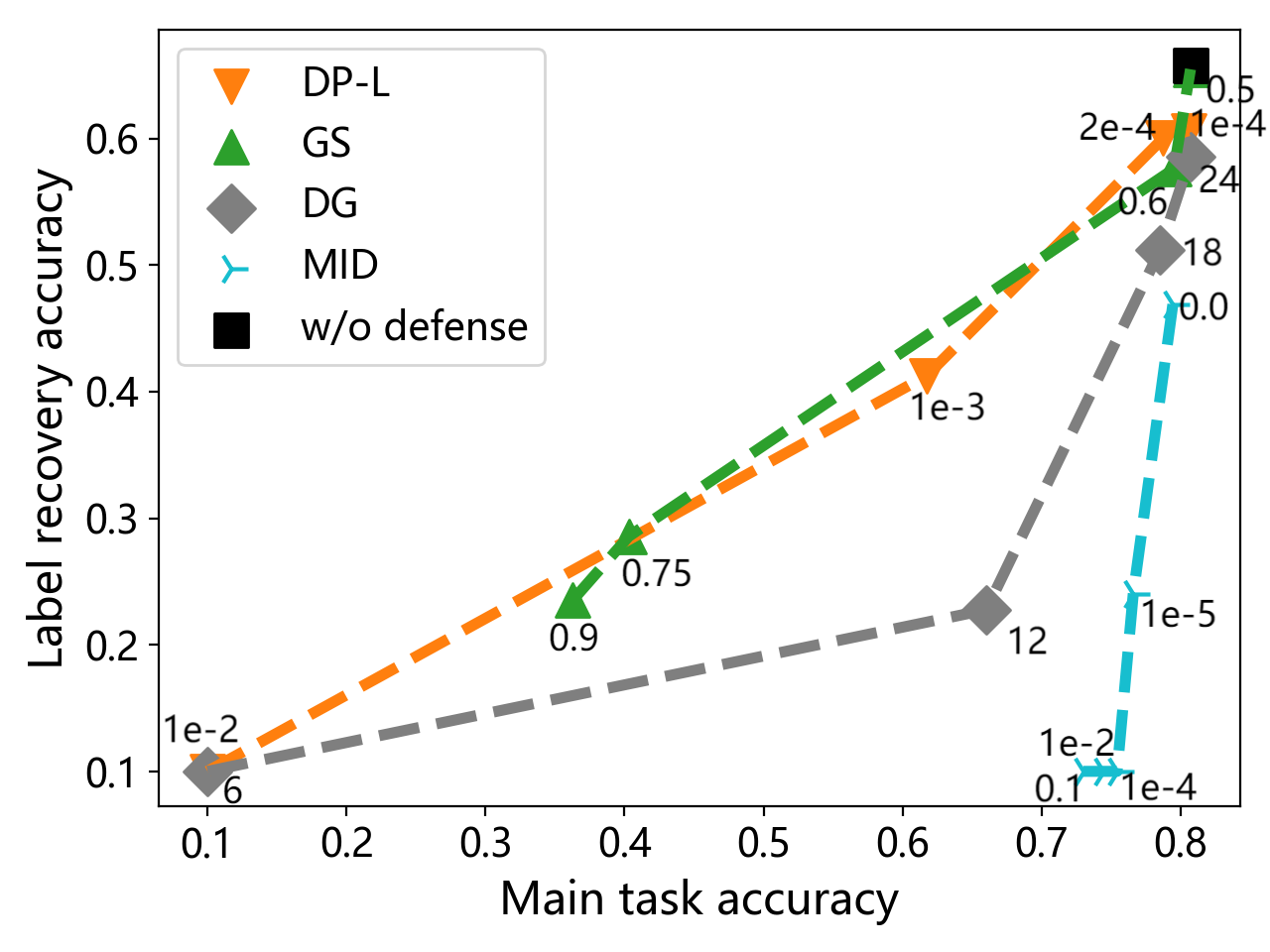}
    \caption{CIFAR10 AMC-10}
    \label{fig:amc_10_cifar10}
  \end{subfigure}

  \begin{subfigure}{0.49\linewidth}
    \includegraphics[width=1\linewidth]{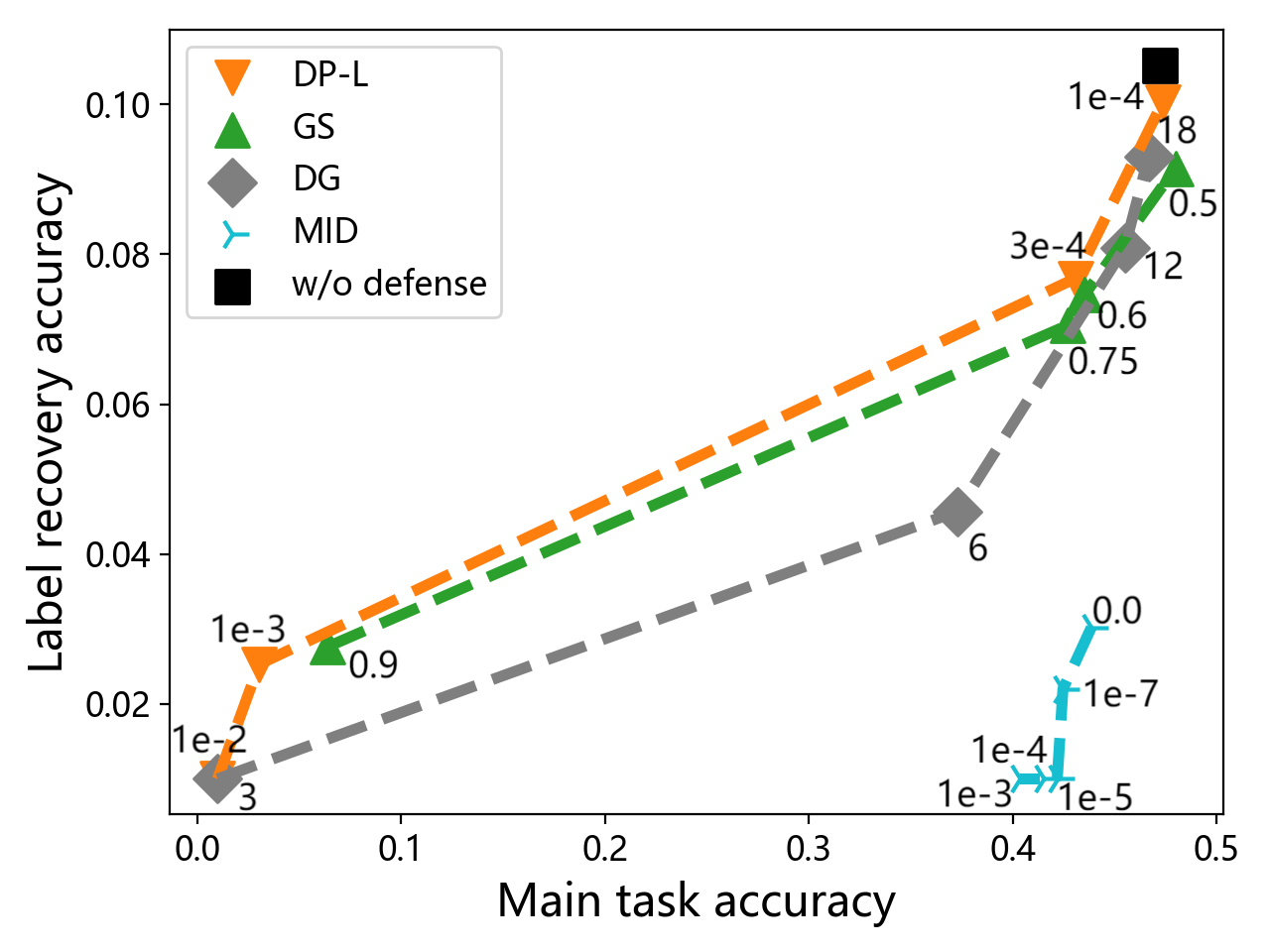}
    \caption{CIFAR100 PMC-100}
    \label{fig:pmc_100_cifar100}
  \end{subfigure}
  \hfill
  \begin{subfigure}{0.49\linewidth}
    \includegraphics[width=1\linewidth]{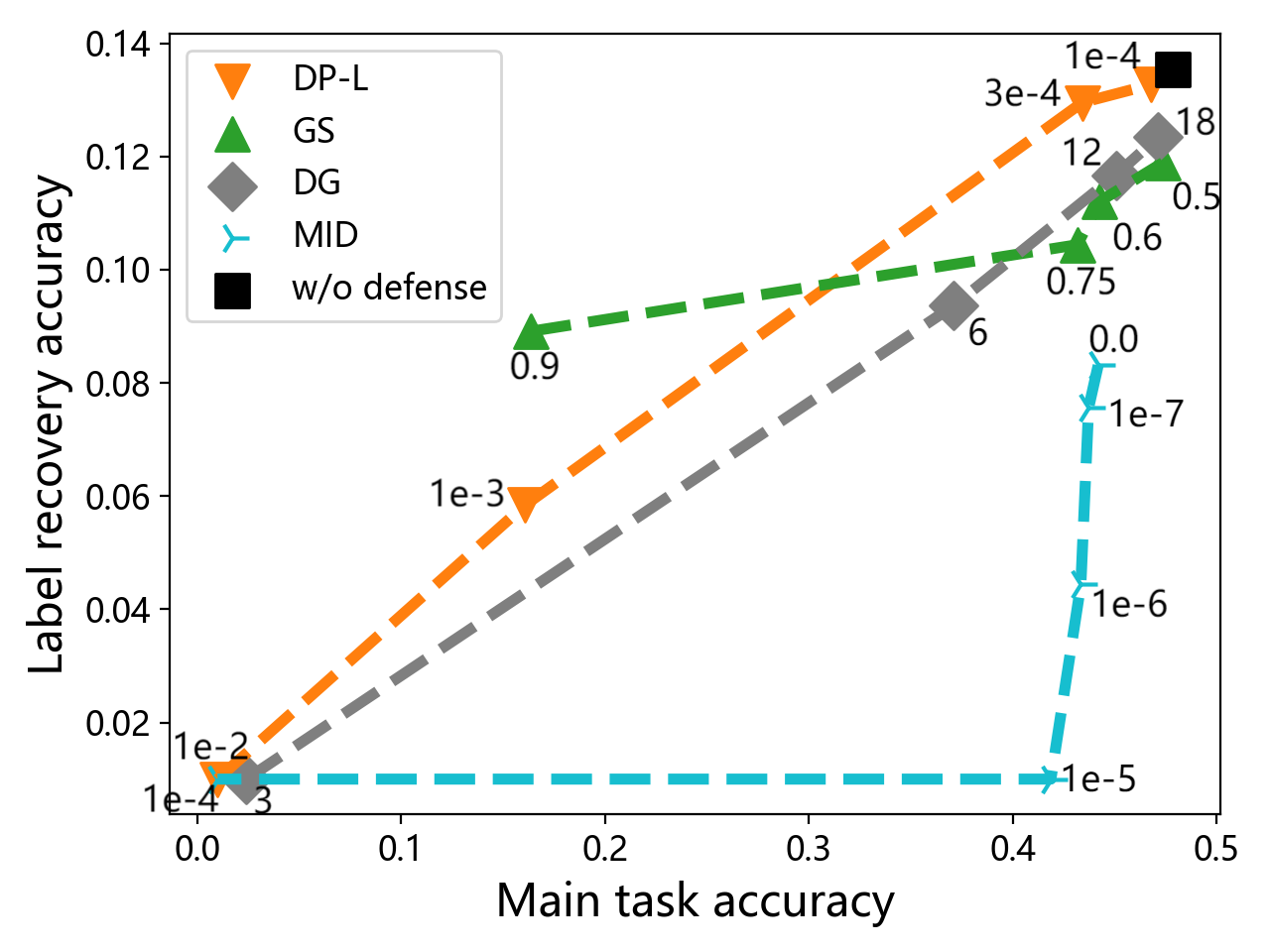}
    \caption{CIFAR100 AMC-100}
    \label{fig:amc_100_cifar100}
  \end{subfigure}
  \caption{Comparison of various kinds of defense methods on passive model completion attack (PMC) and active model completion attack (AMC) using CIFAR10 and CIFAR100 datasets. The number after PMC and AMC is the number of total auxiliary labeled data used in the experiment.} 
  \label{fig:mc_attacks_defenses}
\end{figure}

For our MID defense, we evaluate different hyper-parameters $\lambda$ ranging from $0.0$ to $1e^4$. Note when $\lambda=0.0$, the MID defense is degraded to an encoder-decoder neural network, which is still effective to defend certain gradient-based attacks due to the modification of the local model structure. Comparing it with experimental results of MID with $\lambda > 0$, we can see the effectiveness of generating a variational information bottleneck rather than adding additional model layers to the VFL system.

\subsection{Evaluation Metrics} \label{subsection:evaluation_metrics}

To evaluate different defense methods, we mainly put two metrics in the same figure: attack success rate (y-axis) and main task utility (x-axis). A defense method is considered superior if the attack success rate is lower at the same level of main task utility, thus appearing on the bottom right of the figure. The definition for attack success rate varies slightly for different tasks. For \textbf{\textit{label inference attacks}}, we use the ratio of the correctly recovered labels; for \textbf{\textit{targeted backdoor attack}}, we use backdoor accuracy, the ratio of triggered backdoor samples that are predicted as target class; for \textbf{\textit{non-targeted backdoor attacks}}, we use the drop of main task accuracy on attacked samples; for \textbf{\textit{feature reconstruction attack}}, we use Peak Signal-to-Noise Ratio (PSNR) that is widely utilized for assessing the quality of images \cite{zhu2019deep,jin2021cafe}, where a low PSNR value indicates a high ratio of noise and a low success rate.

\begin{figure} [!tb]
  \centering
  \begin{subfigure}{0.49\linewidth}
    \includegraphics[width=1\linewidth]{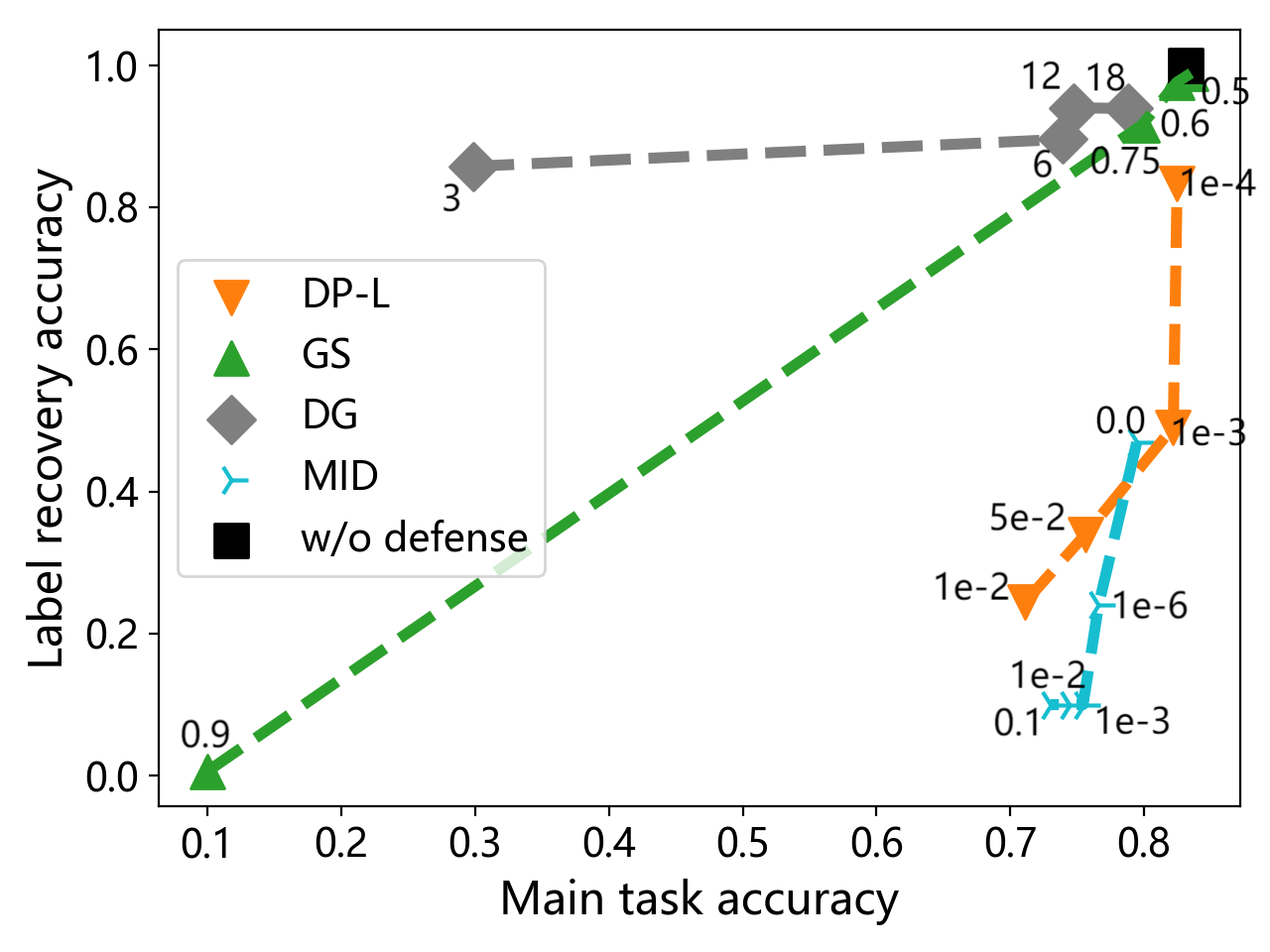}
    \caption{CIFAR10 DLI}
    \label{fig:dli_cifar10}
  \end{subfigure}
  \hfill
  \begin{subfigure}{0.49\linewidth}
    \includegraphics[width=1\linewidth]{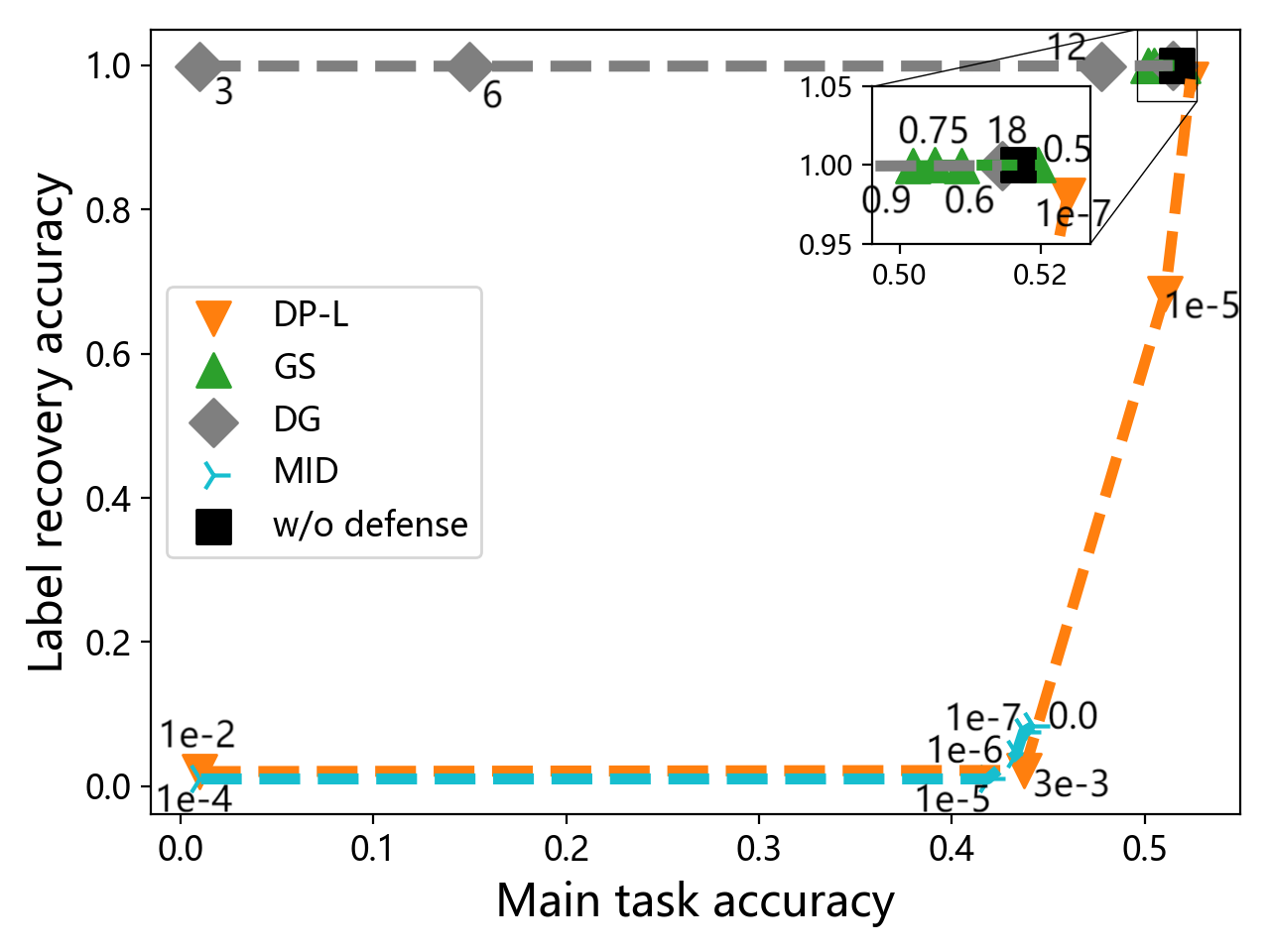}
    \caption{CIFAR100 DLI}
    \label{fig:dli_cifar100}
  \end{subfigure}
  \begin{subfigure}{0.49\linewidth}
    \includegraphics[width=1\linewidth]{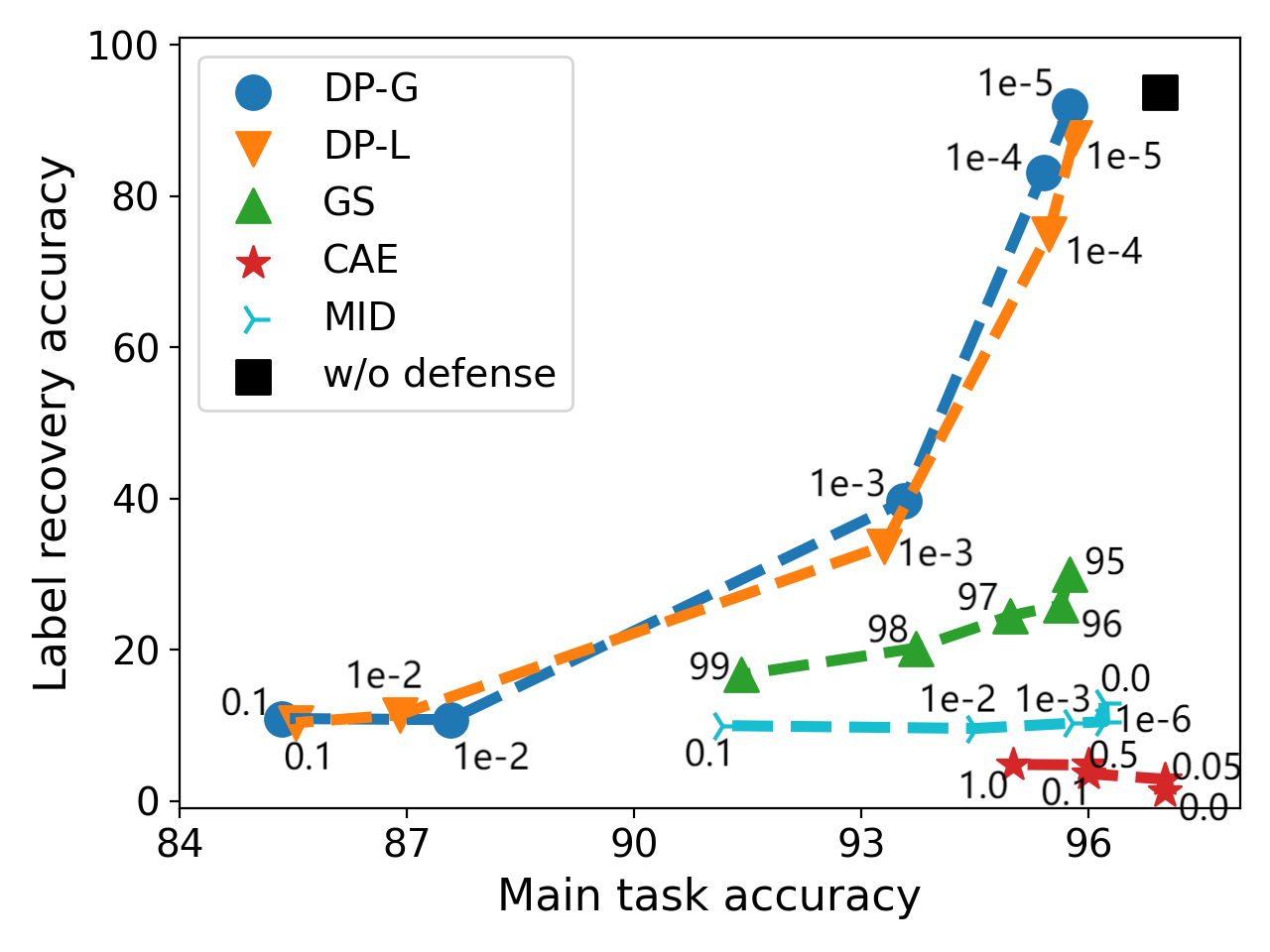}
    \caption{MNIST BLI}
    \label{fig:bli_multi_mnist}
  \end{subfigure}
  \hfill
  \begin{subfigure}{0.49\linewidth}
    \includegraphics[width=1\linewidth]{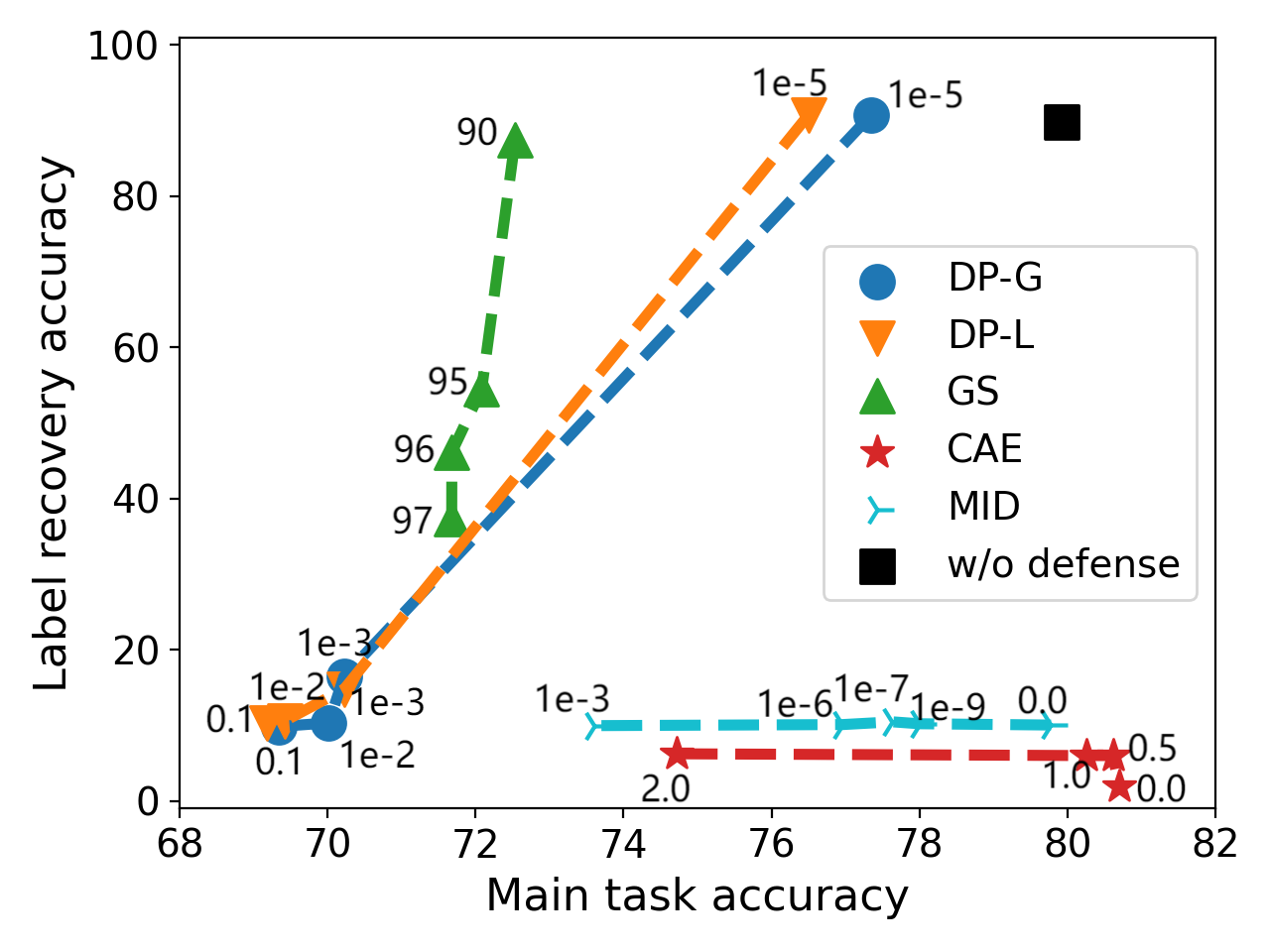}
    \caption{CIFAR10 BLI}
    \label{fig:bli_multi_cifar10}
  \end{subfigure}

  \caption{Comparison of various kinds of defense methods against direct label inference attack (DLI) and batch-level label inference attack (BLI) on $3$ different datasets.}
  \label{fig:label_inference_attacks_defenses}
\end{figure}

\subsection{Defending Against Label Inference Attacks} \label{section:exp_mc_defense}
\textit{\textbf{Model Completion Attacks.}}
We compare MID with $3$ other baseline methods following previous work\cite{fu2021label,zou2022defending}: 
DP-L, GS and DG.
The results are shown in \cref{fig:mc_attacks_defenses} and \cref{fig:mc_attacks_defenses_appendix} in the appendix.
From \cref{fig:mc_attacks_defenses,fig:mc_attacks_defenses_appendix}, we can see that all methods exhibit a trade-off between attack accuracy (y-axis) and main task accuracy (x-axis). Increasing defense strength by increasing noise level, sparsification rate or regularization hyper-parameter $\lambda$ in MID will lead to lower attack accuracy and main task accuracy. However, our MID defense outperforms all the other baseline methods with much lower attack accuracy while maintaining a high main task accuracy over a wide range of $\lambda$ values. The experiments demonstrates the effectiveness of MID defense in suppressing the information of true label distribution $Y$ contained in the local model $G^p$ and local model output $H^p$ at passive party. Other defense methods fail to limit the attack accuracy to the same level when maintaining a similar main task accuracy.


\begin{figure}[!tb]
\centering
  \begin{subfigure}{0.49\linewidth}
    \includegraphics[width=1\linewidth]{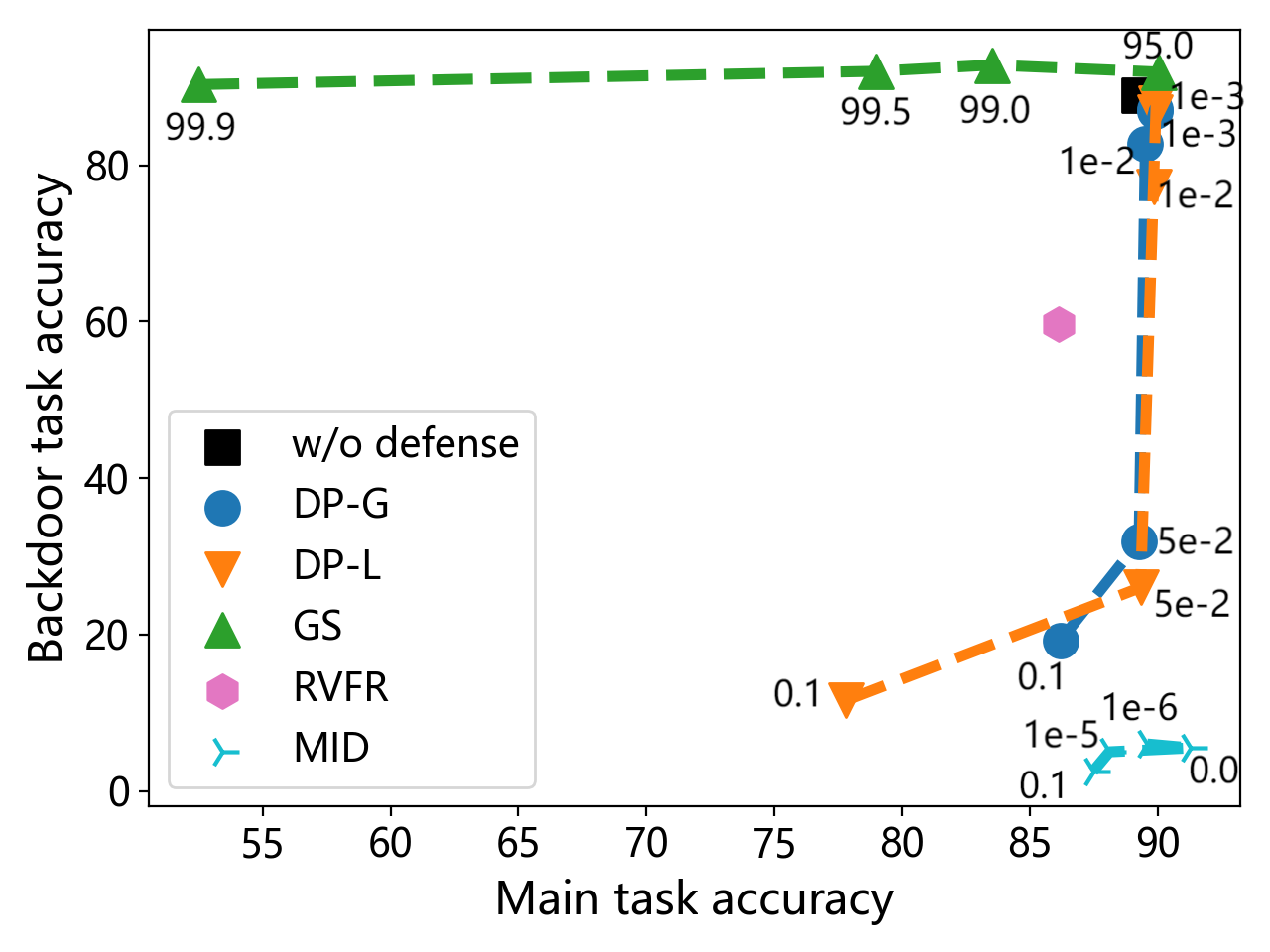}
    \caption{MNIST Targeted}
    \label{fig:backdoor_mnist_2party}
  \end{subfigure}
  \hfill
  \begin{subfigure}{0.49\linewidth}
    \includegraphics[width=1\linewidth]{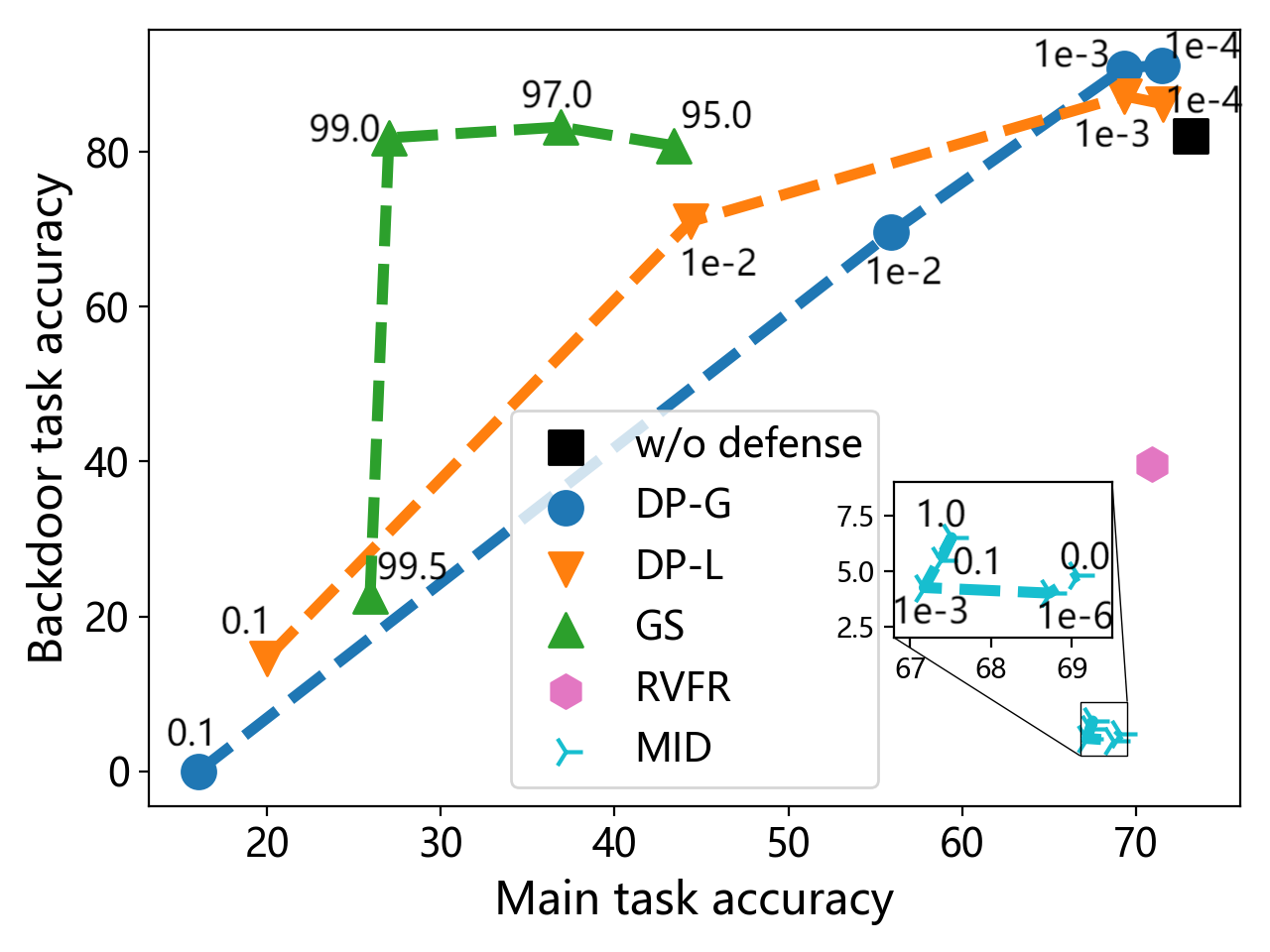}
    \caption{CIFAR10 Targeted}
    \label{fig:backdoor_cifar10_2party}
  \end{subfigure}
  

  \begin{subfigure}{0.49\linewidth}
    \includegraphics[width=1\linewidth]{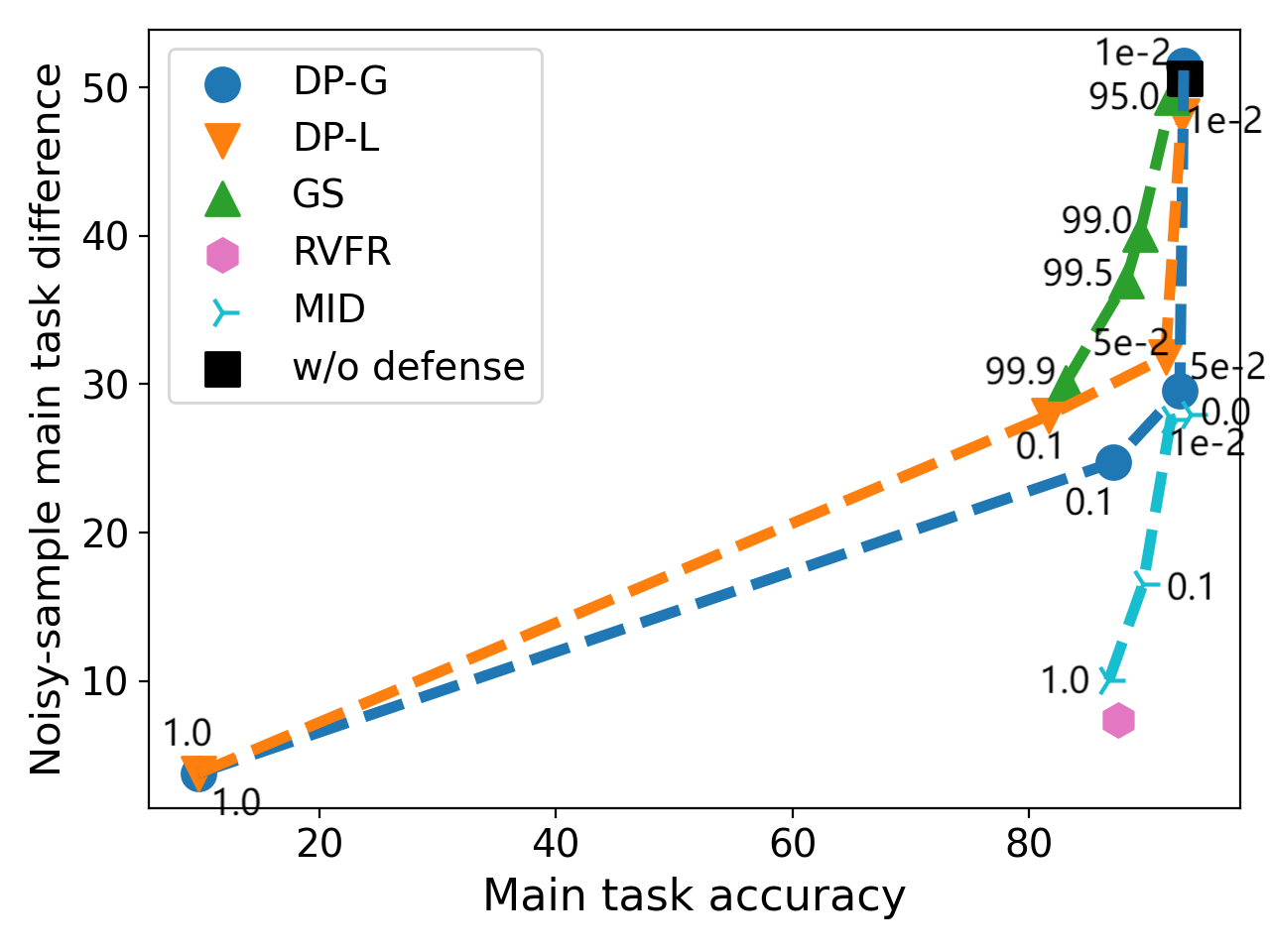}
    \caption{MNIST Noisy-sample }
    \label{fig:noisysample_mnist}
  \end{subfigure}
  \hfill
  \begin{subfigure}{0.49\linewidth}
    \includegraphics[width=1\linewidth]{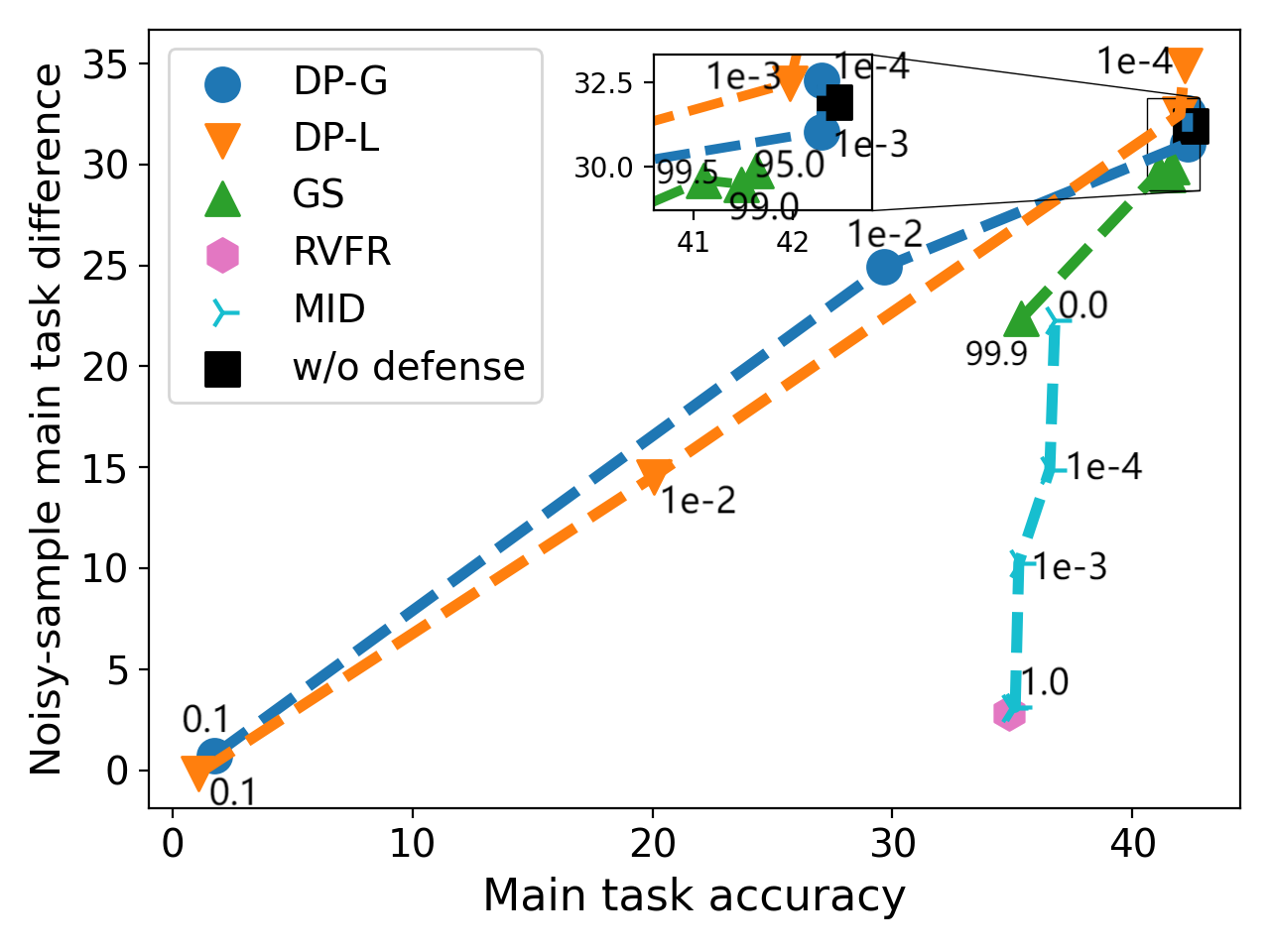}
    \caption{CIFAR100 Noisy-sample}
    \label{fig:noisysample_cifar100}
  \end{subfigure}
  
  \begin{subfigure}{0.49\linewidth}
    \includegraphics[width=1\linewidth]{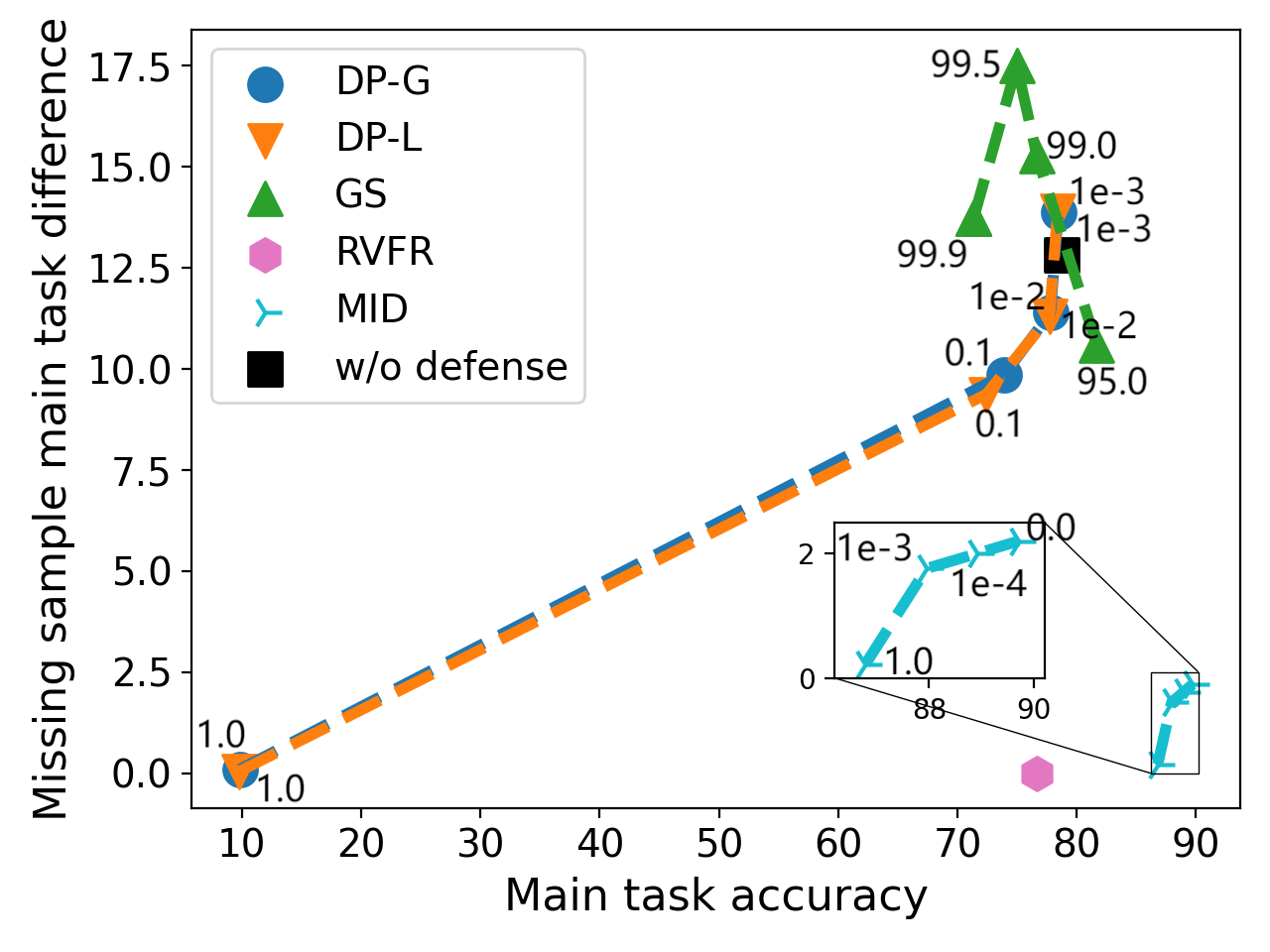}
    \caption{MNIST Missing}
    \label{fig:missing_mnist}
  \end{subfigure}
  \hfill
  \begin{subfigure}{0.49\linewidth}
    \includegraphics[width=1\linewidth]{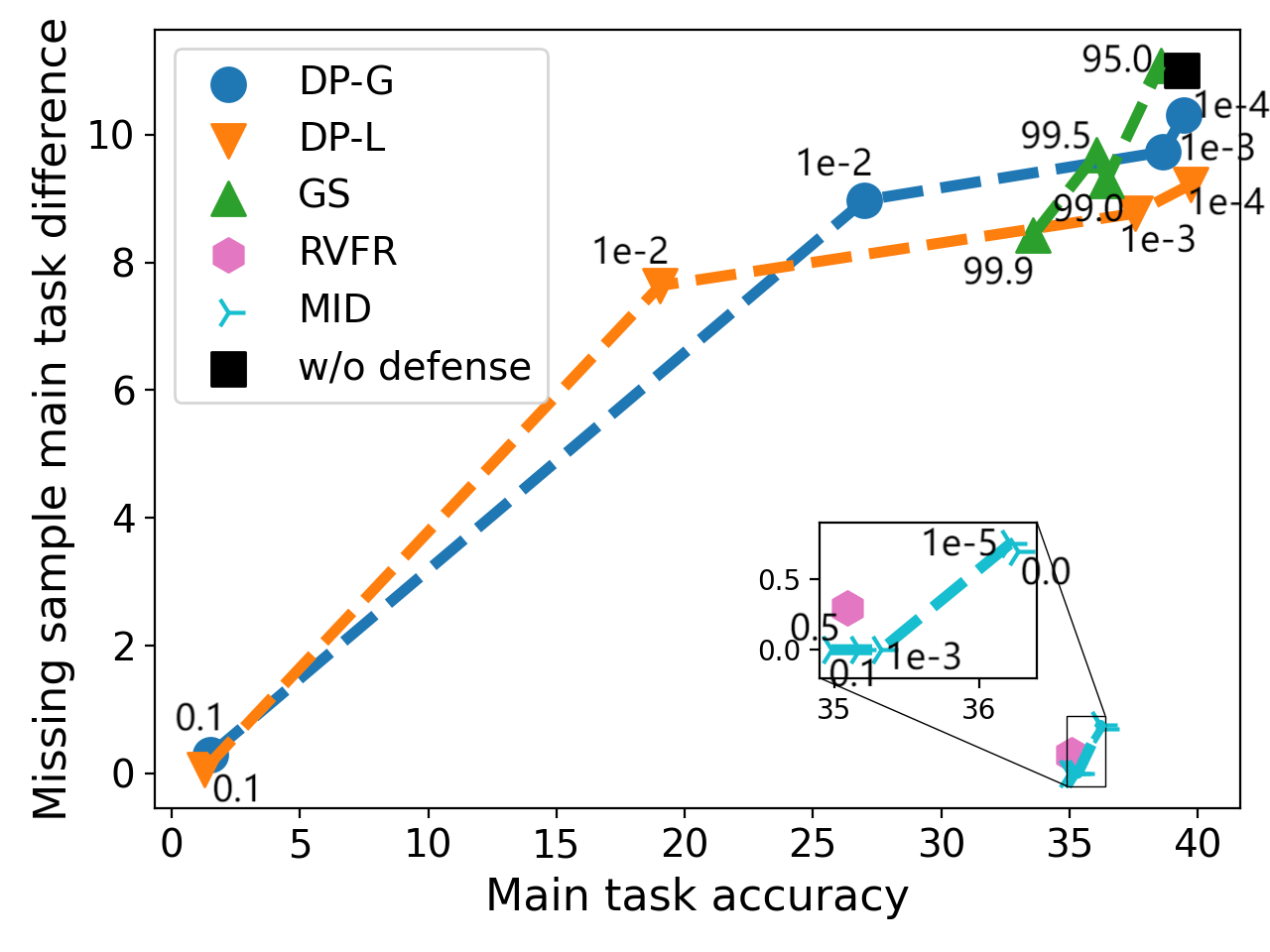}
    \caption{CIFAR100 Missing}
    \label{fig:missing_cifar100}
  \end{subfigure}

  \caption{Comparison of various kinds of defense methods against targeted backdoor attack, namely label replacement backdoor attack, and non-targeted backdoor attacks including noisy-sample backdoor attack and missing backdoor attack on $3$ different datasets.}
    \label{fig:backdoor_attacks_defenses}
\end{figure}

\begin{figure*} [!htb]
  \centering
  \begin{subfigure}{0.32\linewidth}
    \includegraphics[width=1\linewidth]{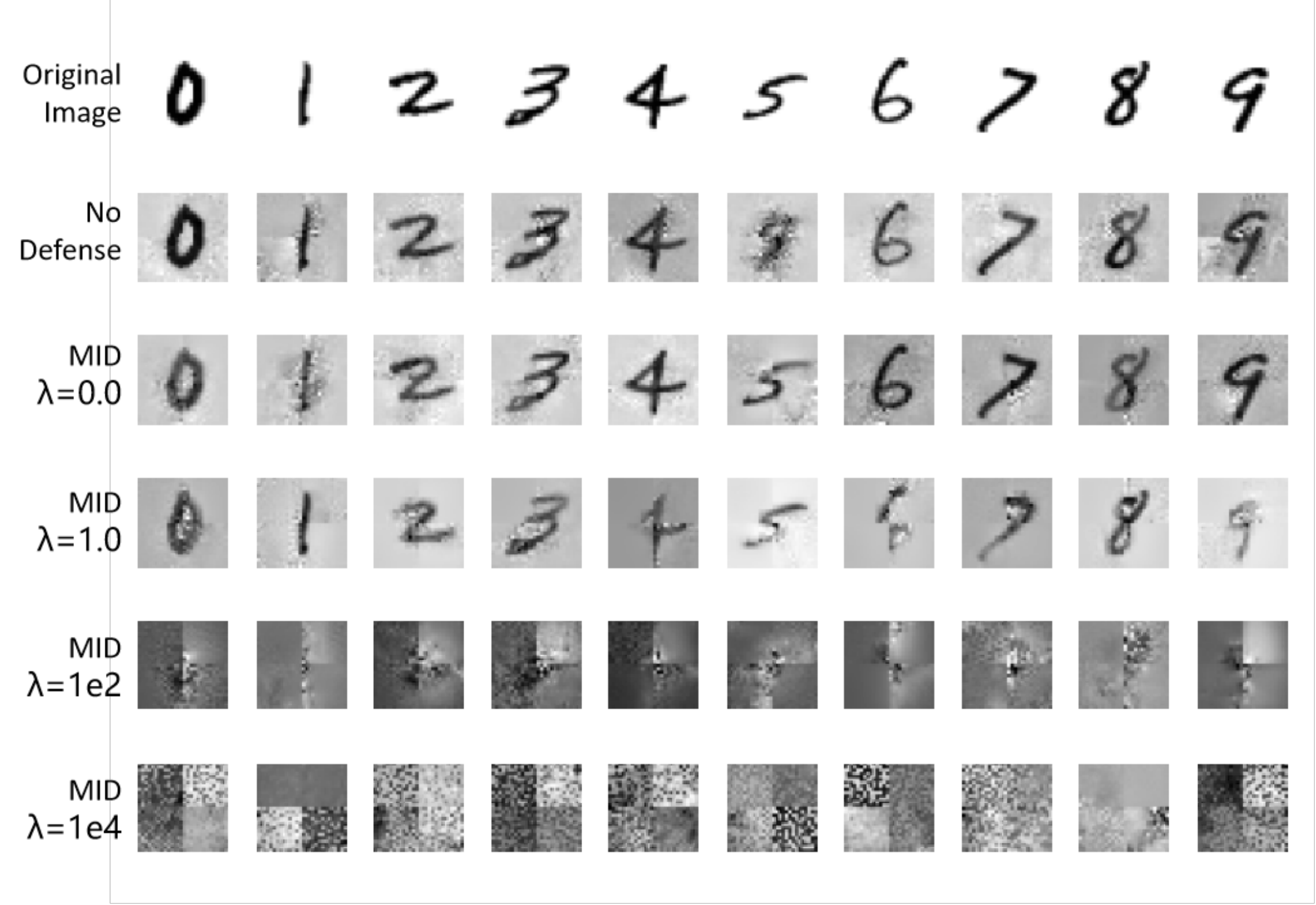}
    \caption{MNIST}
    \label{fig:cafe_mnist}
  \end{subfigure}
  \begin{subfigure}{0.32\linewidth}
    \includegraphics[width=1\linewidth]{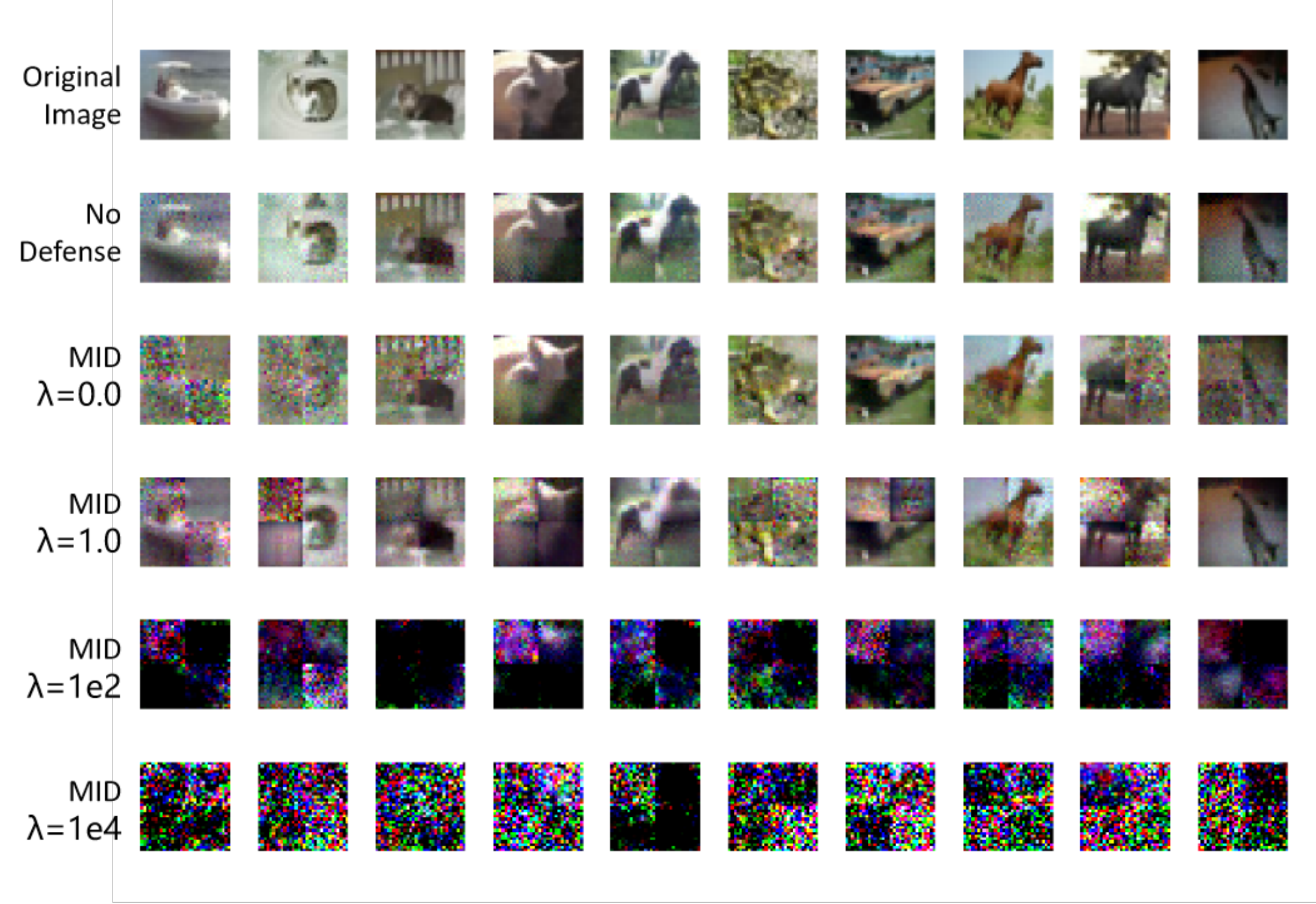}
    \caption{CIFAR10}
  \label{fig:cafe_cifar10}
  \end{subfigure}
  \begin{subfigure}{0.32\linewidth}
    \includegraphics[width=1\linewidth]{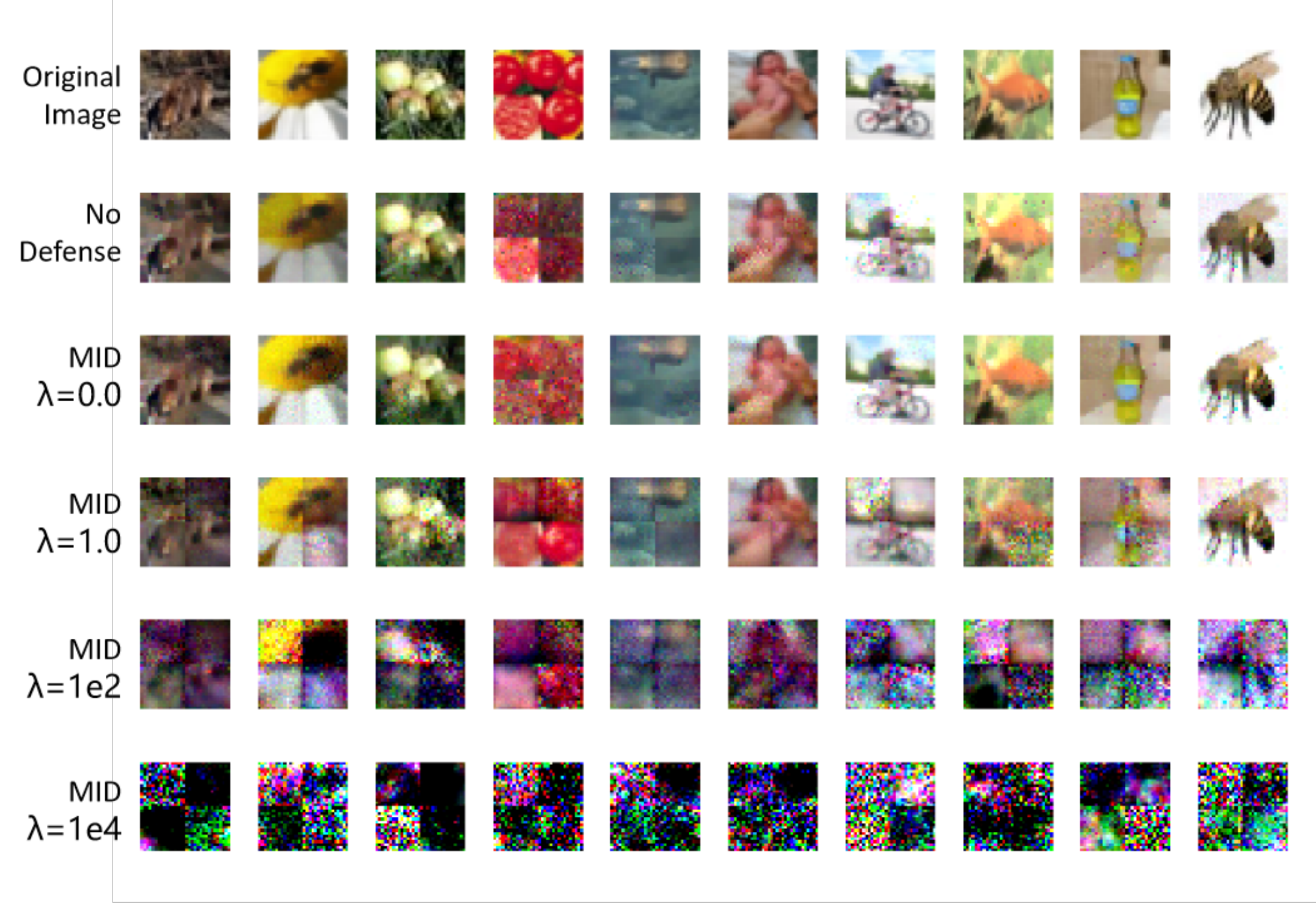}
    \caption{CIFAR100}
  \label{fig:cafe_cifar100}
  \end{subfigure}
  \hfill
  \caption{Effectiveness of MID against CAFE at various $\lambda$.}
  \label{fig:cafe}
\end{figure*}

\textit{\textbf{Sample-level Label Inference Attacks.}} Direct label inference attack (DLI) and direction scoring attack (DS) are $2$ typical types of sample-level label inference attack. We first evaluate MID with $3$ other baseline methods, DP-L, GS and DG, against DLI attack. Defense results are shown in \cref{fig:dli_cifar10,fig:dli_cifar100}. We can see that MID outperforms most of the baseline methods. 
Since DS attack can only recover binary label, aside from DP-G, DP-L and GS, we also compare MID with MARVELL, which is specifically designed for defending DS\cite{li2021label} attack. Results are shown in \cref{fig:ds_defenses_appendix} in the appendix. We can see that MID results in the same level of attack accuracy compared with DP-G, DP-L and GS with a slightly lower main task accuracy than MARVELL. 

\textit{\textbf{Batch-level Label Inference Attacks.}}
We evaluate MID and other defending methods against BLI attack with MNIST and CIFAR10 dataset, and results are shown in \cref{fig:bli_multi_mnist,fig:bli_multi_cifar10}. 
It's clear that, MID performs better than DP-G, DP-L and GS, the $3$ commonly used defending methods under VFL scenario, with a much lower attack accuracy while maintaining the same level main task accuracy. Notice that CAE, a specific defense method designed for BLI attack in which real labels are disguised with soft fake labels, achieves the same level of main task accuracy with a slightly lower attack accuracy compared to MID.


\subsection{Defending Against Backdoor Attacks} \label{subsec:defend_backdoor_result}


The results for backdoor attacks are shown in \cref{fig:backdoor_attacks_defenses}. From \cref{fig:backdoor_mnist_2party,fig:backdoor_cifar10_2party}, we can see that MID is the most effective defense method among all the $5$ defending methods we evaluated (MID, DP-G, DP-L, GS and RVFR), as it achieves a much lower backdoor success rate at a high main task accuracy for \textit{\textbf{targeted backdoor attack}}. 
For \textit{\textbf{non-targeted backdoor attacks}}, MID is also the most effective defending method, especially for missing attack (see \cref{fig:missing_mnist,fig:missing_cifar100}). For noisy sample attack, RVFR is slightly better than or comparable to MID (\cref{fig:noisysample_mnist,fig:noisysample_cifar100}). Notice that the point with $\lambda=0.0$ appears closer to the bottom of \cref{fig:backdoor_mnist_2party,fig:backdoor_cifar10_2party,fig:missing_mnist,fig:missing_cifar100},  due to the fact that targeted backdoor attack and missing attack are more vulnerable to the changes in the model settings, i.e., when $\mathcal{M_{VIB}}$ is added, resulting in a low attack accuracy even without information regularization. 


\subsection{Defending Against Feature Reconstruction Attack}

As the attacker is the active party under this setting, MID is applied at the passive party like shown in \cref{fig:passive_mid_implementation}.


Results of reconstruction images are shown in \cref{fig:cafe} and \cref{tab:cafe_cifar10}. More results are listed in \cref{tab:cafe_cifar10_appendix} in the appendix. We can see that CAFE successfully recovers original data using gradients and local models. However MID and DP-G can both successfully prevent the attacker from successfully recover data features while MID can maintain a high main task accuracy at the same time.
With the increase of $\lambda$ in MID, the model becomes more robust against feature reconstruction attack since less information can be recovered within the same number of iterations, both visually and quantitatively indicated by a lower PSNR value, while the model utility is just slightly harmed (see \cref{tab:cafe_cifar10}). 
Compared with DP-G, MID can simultaneously achieve a lower PSNR value and a much higher main task accuracy as shown in \cref{tab:cafe_cifar10}, indicating a better defense ability against reconstruction attacks. 

\begin{table}
  \centering
  \resizebox{0.9\linewidth}{!}{
  \begin{tabular}{@{}l|cc|cc@{}}
    \toprule
    \multirow{2}*{\shortstack{Defense Method}} & \multicolumn{2}{c|}{CIFAR10} & \multicolumn{2}{c}{CIFAR100}\\
    \cline{2-5}
    \\[-1em]
    ~ & \shortstack{PSNR\\Value} & \shortstack{Main\\ACC} & \shortstack{PSNR\\Value} & \shortstack{Main\\ACC}\\
    \midrule
    No defense & 21.4417 & 0.6015 & 20.5476 & 0.3296 \\
    MID, $\lambda=0.0$ & 20.2628 & 0.5956 & 20.4584 & 0.3281\\
    MID, $\lambda=1.0$ & 18.6929 & 0.5920 & 18.9796 & 0.3235\\
    MID, $\lambda=100.0$ & 8.6667 & 0.5881 & 11.4972 & 0.3213\\
    MID, $\lambda=10000.0$ & 6.1831 & 0.5844 & 6.1711 & 0.3209\\
    DP-G, $\epsilon=0.1$ & 7.1257 & 0.2754 & 6.6617 & 0.0525\\
    \bottomrule
  \end{tabular}
  }
  \caption{PSNR value for recovered data and main task accuracy of CAFE for CIFAR10 and CIFAR100 datasets.}
  \label{tab:cafe_cifar10}
\end{table}

\section{Conclusion}
In this paper, we introduce a novel general defense method MID which is able to defend against various kinds of label inference attacks, backdoor attacks and feature reconstruction attacks under VFL scenario. We provide theoretical analysis and comprehensive experimental evaluations to testify the effectiveness of MID compared to existing defense methods. We believe this work will shed light on future research directions towards improving privacy and robustness of VFL systems. 
{\small
\bibliographystyle{ieee_fullname}
\bibliography{egbib}
}

\clearpage

\appendix
\setcounter{page}{1}
\label{main_appendix}

\section*{Appendix}

\section{Algorithm of MID adopted by passive party}
We describe how MID is applied at passive party in detail in \cref{alg:VFL_mid_passive}. 

\begin{algorithm}[H] 
\caption{A VFL framework with MID (at passive party)}
\textbf{Input}: Learning rate $\eta$; MID hyper-parameter $\lambda$\\
\textbf{Output}: Model parameters $\theta_1,\theta_2,\dots,\theta_K$

\begin{algorithmic}[1] \label{alg:VFL_mid_passive}
\STATE Party 1,2,\dots,$K$, initialize $\theta_1$, $\theta_2$, ... $\theta_K$; \\
\FOR{each iteration j=1,2, ...}
\STATE Randomly sample  $S \subset [N]$;
\FOR{each passive party $k$ ($\ne K$) in parallel}
\STATE Computes $\{H_i^k\}_{i \in S}$;
\STATE Applies MID to generate $Z_i^k=\mathcal{M_{MID}}^k(H_i^k)$;
\STATE Sends $\{Z_i^k\}_{i \in S}$ to party $K$;
\ENDFOR
\STATE Active party $K$ computes $\{H_i^K\}_{i \in S}$;
\STATE Active party computes loss $\ell$ using \cref{eq:mid_loss_multi}; 
\STATE Active party sends $\{\frac{\partial \ell}{\partial Z_i}\}_{i \in S}$ to all other parties;
\FOR{each party k=1,2,\dots,K in parallel}
\STATE Passive party $k(\ne K)$ computes $\{\frac{\partial \ell}{\partial Z_i} \leftarrow \frac{\partial \ell}{\partial Z_i}+\frac{\partial I(H^k,Z^k)}{\partial Z_i^k}\}_{i \in S}$ and $\nabla_k\ell= \frac{\partial \ell}{\partial Z_i}\frac{\partial Z_i^k}{\partial \theta_k}$;
\STATE Active party $K$ computes $\nabla_K\ell= \frac{\partial \ell}{\partial H_i}\frac{\partial H_i^K}{\partial \theta_K}$;
\STATE Each party updates $\theta^{j+1}_k = \theta^{j}_k - \eta \nabla_k\ell$;
\ENDFOR
\ENDFOR
\end{algorithmic}
\end{algorithm}

The main difference between this algorithm and \cref{alg:VFL_mid_active} is that in this algorithm, $\mathcal{M_{MID}}^k$ is now kept at each passive party instead of the active party in \cref{alg:VFL_mid_active}.

\section{Proof for \cref{theorem:mid_backdoor}}
\begin{proof}

From the relation of mutual information to entropy and conditional entropy, i.e. $I(X,Y)= H(X)-H(X|Y)$, we have:
\begin{equation*}
\begin{split}
    &|I(Y,T)-I(Y,T^{\prime})| \\ &= |H(T)-H(T|Y)-H(T^{\prime})+H(T^{\prime}|Y)| \\ &= |[H(T)-H(T^{\prime})]-[H(T|Y)-H(T^{\prime}|Y)]| \\ &\leq |H(T|Y)-H(T^{\prime}|Y)|+|H(T)-H(T^{\prime})|\\
\end{split}
\end{equation*}
In the following, we will show that $|H(T|Y)-H(T^{\prime}|Y)|$ and $|H(T)-H(T^{\prime})|$ each has an upper bound. 

Following Theorem 3.2 in \cite{wang2020infobert}, $|H(T|Y)-H(T^{\prime}|Y)|$ has an upper bound:

\begin{equation}\label{eq:upper_bound_HTY_appendix}
\begin{split}
    |H(T|Y)-H(T^{\prime}|Y)| &\leq B_2 \log\frac{1}{B_2} {\mathcal{|T|}}^{1/2}(I(H^p,T))^{1/2}\\
    &+ B_2 {\mathcal{|T|}}^{3/4}(I(H^p,T))^{1/4}\\ 
    &+ B_4 \log\frac{1}{B_4} {\mathcal{|T|}}^{1/2}(I({H^p}^{\prime},T^{\prime}))^{1/2}\\
    &+ B_4 {\mathcal{|T|}}^{3/4}(I({H^p}^{\prime},T^{\prime}))^{1/4}
\end{split}
\end{equation}
This upper bound is symmetric to $T$ and $T^{\prime}$ and is positively correlated to $I(H^p,T)$ and $I({H^p}^{\prime},T^{\prime})$ respectively. If we define $B_1 \triangleq B_2 \log\frac{1}{B_2}$ and $B_3 \triangleq B_4 \log\frac{1}{B_4}$, then \cref{eq:upper_bound_HTY_appendix} has the form that is the same to the first $4$ items of the right side of \cref{eq:infobert_theorem}. Notice that, the four coefficients $B_1, B_2, B_3, B_4$ and $|\mathcal{T}|$, the size of the finite set of possible values of $T$, are all independent of $H^p$ and $T$.

Moreover, $|H(T)-H(T^{\prime})|$ can be bounded with a constant value. If $t\in \mathcal{T}, t^{\prime} \in \mathcal{T^{\prime}}$ satisfy $||t-t^{\prime}||_2\leq\epsilon$, then we refer to $t^{\prime}$ as an $\epsilon$-bounded modified representation of $t$. If we denote the number of the $\epsilon$-bounded modified representation $t^{\prime}$ around $t$ as $M(t)$, then following Equation $(82)$ in \cite{wang2020infobert}, we have:

\begin{equation}\label{eq:bound_HT_appendix}
\begin{split}
   |H(T)-H(T^{\prime})| &\leq |\sum_{t \in \mathcal{T}} p(t) \log M(t)| \\ &\leq |\sum_{t \in \mathcal{T}} p(t) \log M| \\ &= |\log M|
\end{split}
\end{equation}
where $M=\sup_{t\in \mathcal{T}}{M(t)}$. This means, $|H(T)-H(T^{\prime})|$ can be bounded by a value independent to $H^p, T$ and $\epsilon$. 

Summing up \cref{eq:upper_bound_HTY_appendix,eq:bound_HT_appendix}, we can get \cref{eq:infobert_theorem}. And \cref{eq:infobert_theorem} shows that \cref{eq:backdoor_defense_target} can be achieved by achieving $\min I(H^p,T)$.

\end{proof}

\section{Additional Experimental Results}

\subsection{Defending Against Label Inference Attacks}

\begin{figure} [!tb]
  \centering
  \begin{subfigure}{0.49\linewidth}
    \includegraphics[width=1\linewidth]{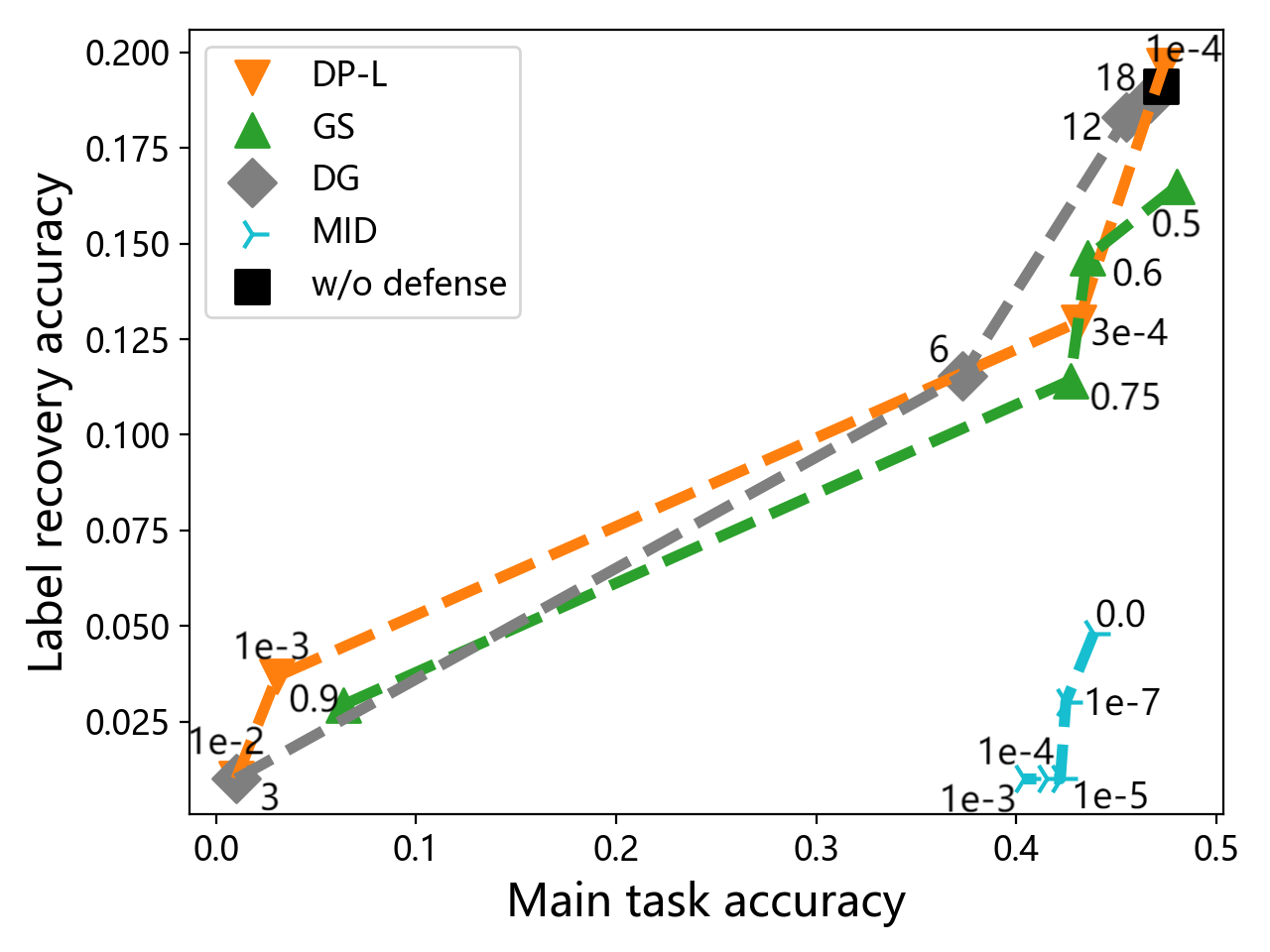}
    \caption{CIFAR100 PMC-400}
    \label{fig:pmc_400_cifar100}
  \end{subfigure}
  \hfill
  \begin{subfigure}{0.49\linewidth}
    \includegraphics[width=1\linewidth]{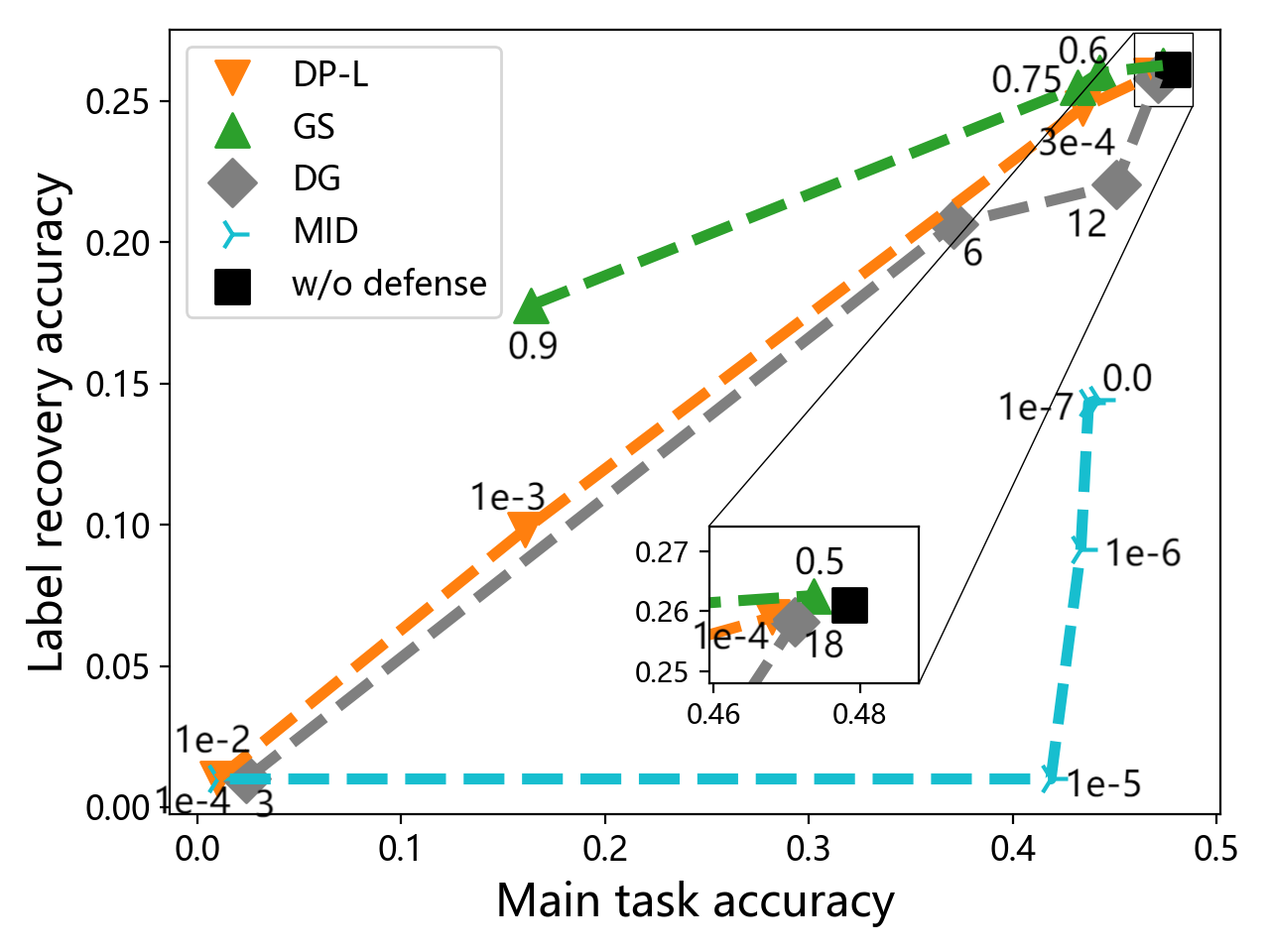}
    \caption{CIFAR100 AMC-400}
    \label{fig:amc_400_cifar100}
  \end{subfigure}
  
  \caption{Comparison of various kinds of defense methods on passive and active model completion attack (PMC, AMC) using CIFAR100 dataset. The number after PMC and AMC is the number of total auxiliary labeled data used in the experiment.} 
  \label{fig:mc_attacks_defenses_appendix}
\end{figure}

The results of various defending methods against \textit{\textbf{model completion attacks}}, including \textit{passive model completion attack} (PCM) and \textit{active model completion attack} (ACM), on CIFAR100 dataset with 400 auxiliary labeled data are presented in \cref{fig:mc_attacks_defenses_appendix}. It's clear to see from the figure that MID performs better than the $3$ other baseline methods, including DP-G, DP-L and GS, since a lower recovery accuracy is achieved at the same level of main task utility.

\begin{figure} [!tb]
  \centering
  \begin{subfigure}{0.49\linewidth}
    \includegraphics[width=1\linewidth]{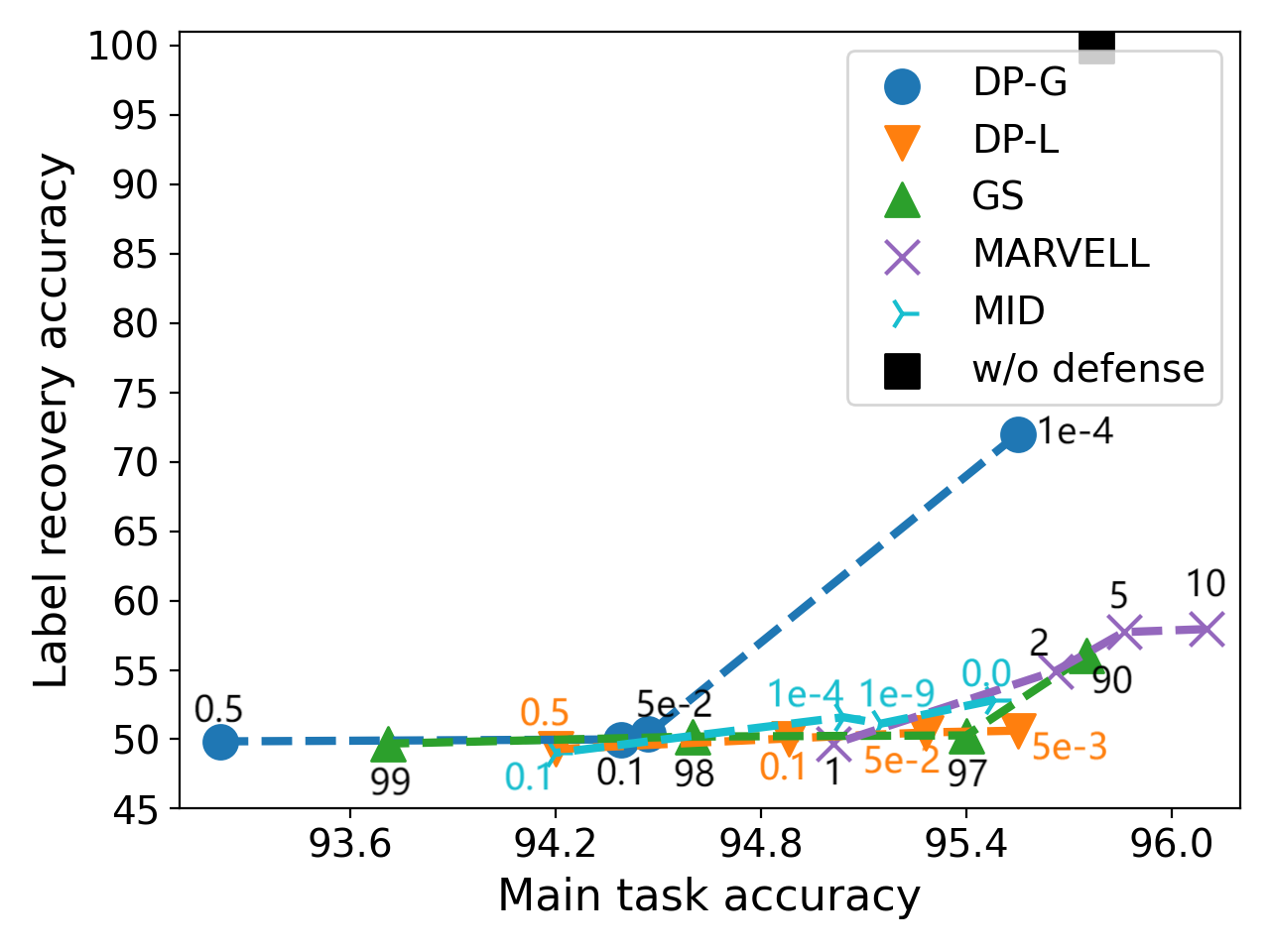}
    \caption{CIFAR10 DS}
    \label{fig:ds_cifar10}
  \end{subfigure}
  \hfill
  \begin{subfigure}{0.49\linewidth}
    \includegraphics[width=1\linewidth]{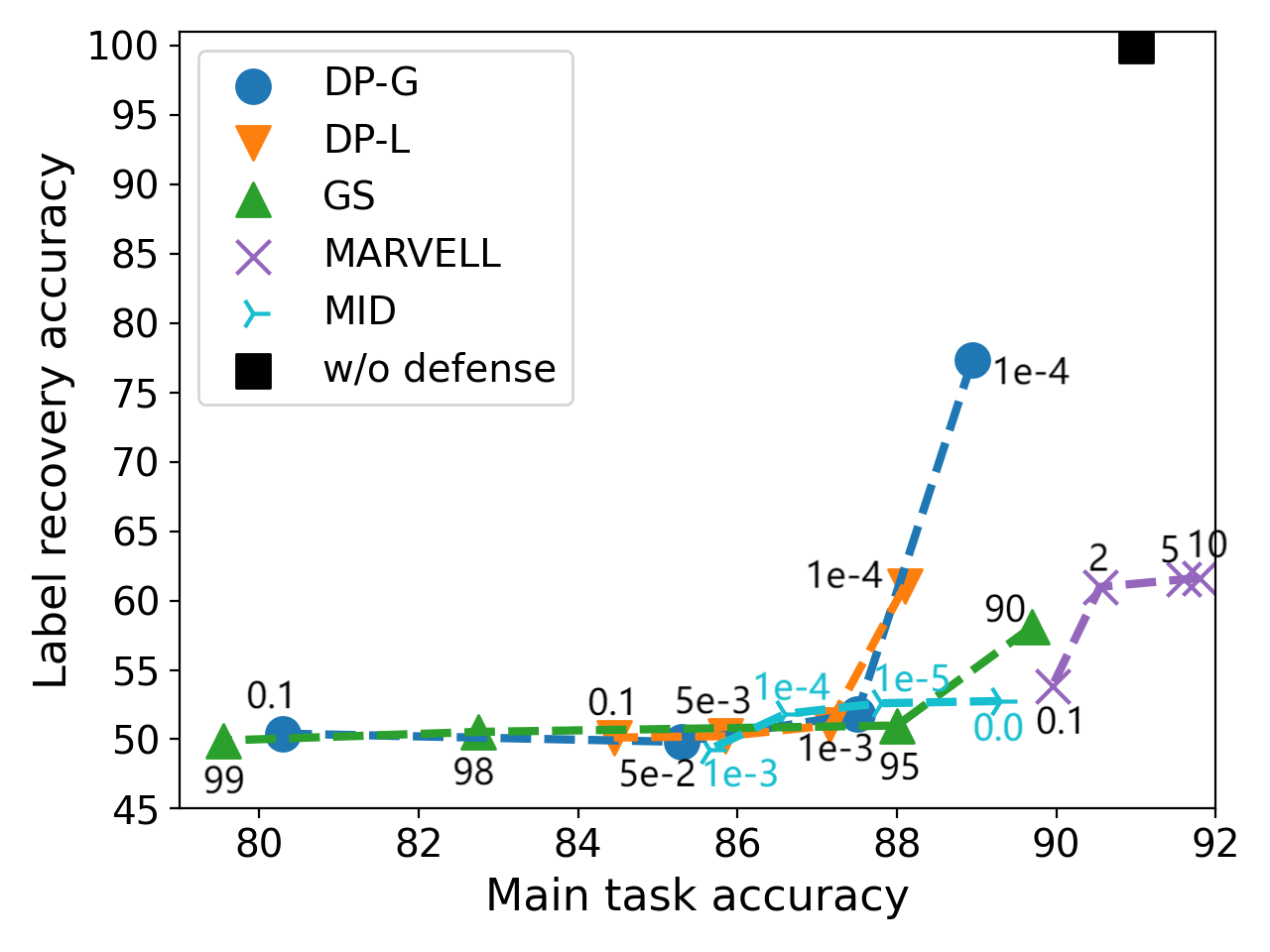}
    \caption{CIFAR100 DS}
    \label{fig:ds_cifar100}
  \end{subfigure}

  \caption{Comparison of various kinds of defense methods against direction scoring attack (DS) on CIFAR10 and CIFAR100 datasets.}
  \label{fig:ds_defenses_appendix}
\end{figure}

For \textit{direction scoring attack} (DS), the defense results are shown in \cref{fig:ds_defenses_appendix}. We can see from \cref{fig:ds_cifar10,fig:ds_cifar100} that all the defense methods can reduce the attack accuracy to a low level with comparable main task accuracy while MARVELL achieves a slightly higher main task accuracy. 

\subsection{Defending Against Backdoor Attacks}

\begin{figure}[!tb]
\centering
  \begin{subfigure}{0.49\linewidth}
    \includegraphics[width=1\linewidth]{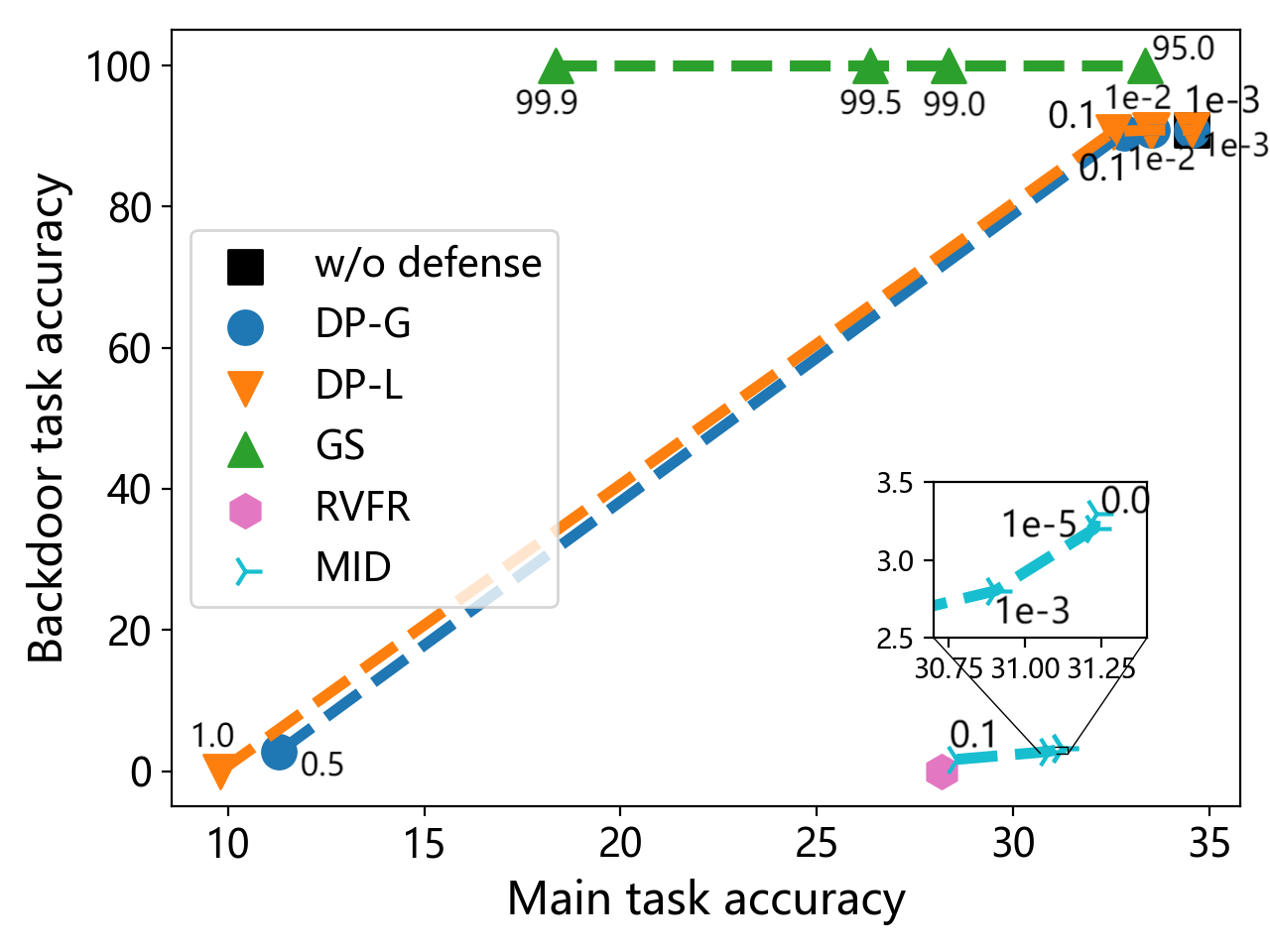}
    \caption{4-party Backdoor MNIST}
    \label{fig:backdoor_mnist_4party}
  \end{subfigure}
  \hfill
  \begin{subfigure}{0.49\linewidth}
    \includegraphics[width=1\linewidth]{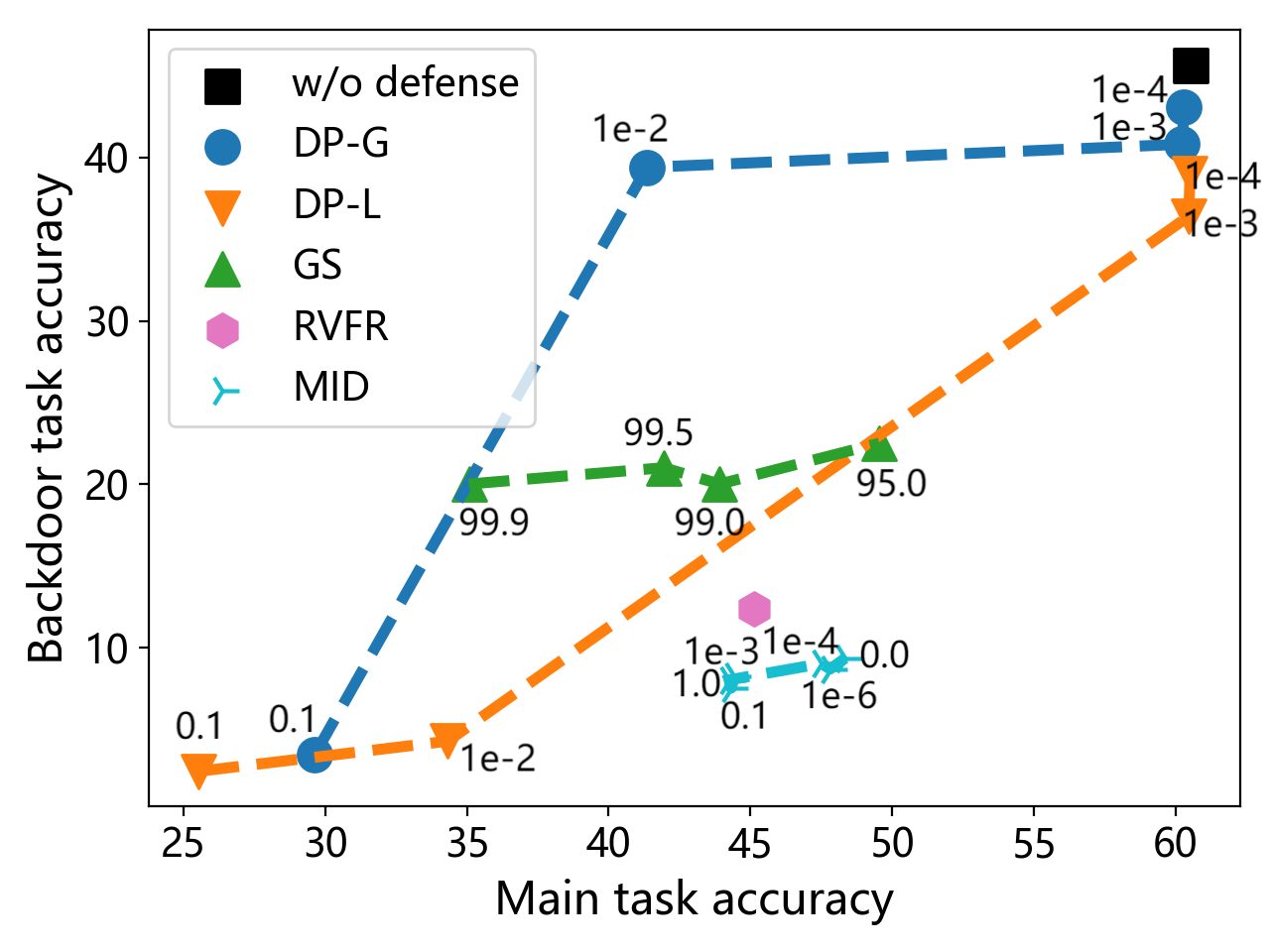}
    \caption{4-party Backdoor CIFAR10}
    \label{fig:backdoor_cifar10_4party}
  \end{subfigure}

  \caption{Comparison of various kinds of defense methods against 4-party targeted backdoor attack on MNIST dataset and CIFAR10 dataset.}
    \label{fig:targeted_backdoor_attacks_defenses_appendix}
\end{figure}

We also conduct \textit{\textbf{targeted backdoor attack}} under 4-party VFL setting. In this setting, the $3$ passive parties cooperate with each other by sharing the same target label $\tau$ and adding local triggers to the same set of triggered samples to launch a gradient replacement backdoor attack. We evaluate MID with the same $4$ other baseline defense mechanisms we use in \cref{subsec:defend_backdoor_result} and the results are presented in \cref{fig:targeted_backdoor_attacks_defenses_appendix}. From this figure, we can see that MID can limit the backdoor accuracy to a much lower level compared to other methods (DP-G, DP-L and GS). Moreover, RVFR, a defense designed for defending against backdoor attacks, achieves a similar defense ability compared with MID.

\subsection{Defending Against Feature Reconstruction Attack}

We present more experimental results of MID and DP-G against \textbf{CAFE} attack in \cref{tab:cafe_cifar10_appendix}. We evaluate MID with hyper-parameter $\lambda$ ranging from $0.0$ to $10000.0$ and DP-G with noise of standard deviation ranging from $0.1$ to $10.0$ following the original work\cite{jin2021cafe}. Results for DP-G exhibit very similar trend, consistent with the original work\cite{jin2021cafe}. For MID, we observe that as $\lambda$ increases, the feature reconstruction quality is worsened, but the main task accuracy is not affected as much as defense with DP-G.

\begin{table}
  \centering
  \resizebox{0.9\linewidth}{!}{
  \begin{tabular}{@{}l|cc|cc@{}}
    \toprule
    \multirow{2}*{\shortstack{Defense Method}} & \multicolumn{2}{c|}{CIFAR10} & \multicolumn{2}{c}{CIFAR100}\\
    \cline{2-5}
    \\[-1em]
    ~ & \shortstack{PSNR\\Value} & \shortstack{Main\\ACC} & \shortstack{PSNR\\Value} & \shortstack{Main\\ACC}\\
    \midrule
    No defense & 21.4417 & 0.6015 & 20.5476 & 0.3296 \\
    MID, $\lambda=0.0$ & 20.2628 & 0.5956 & 20.4584 & 0.3281\\
    MID, $\lambda=0.1$ & 20.0116 & 0.5944 & 19.2968 & 0.3249\\
    MID, $\lambda=1.0$ & 18.6929 & 0.5920 & 18.9796 & 0.3235\\
    MID, $\lambda=10.0$ & 15.4265 & 0.5908 & 16.3231 & 0.3230\\
    MID, $\lambda=100.0$ & 8.6667 & 0.5881 & 11.4972 & 0.3213\\
    MID, $\lambda=1000.0$ & 7.1028 & 0.5873 & 6.9704 & 0.3219\\
    MID, $\lambda=10000.0$ & 6.1831 & 0.5844 & 6.1711 & 0.3209\\
    DP-G, $\epsilon=0.1$ & 7.1257 & 0.2754 & 6.6617 & 0.0525\\
    DP-G, $\epsilon=1.0$ & 7.1126 & 0.2742 & 6.6578 & 0.0497\\
    DP-G, $\epsilon=10.0$ & 7.1174 & 0.2722 & 6.1279 & 0.0489\\
    \bottomrule
  \end{tabular}
  } 
  \caption{PSNR value for recovered data and main task accuracy of CAFE for CIFAR10 and CIFAR100 datasets.} 
  \label{tab:cafe_cifar10_appendix}
\end{table}

\end{document}